\documentclass[jair,twoside,11pt,theapa]{article}
\usepackage{xcolor}
\usepackage[colorlinks=true,linkcolor=blue,citecolor=blue, urlcolor=blue]{hyperref}
\usepackage{jair, theapa, rawfonts}

\ShortHeadings{Domain-Adaptive Trajectory Imitation}
{Solano-Carrillo \& Stoppe}
\firstpageno{1}

\usepackage{microtype}
\usepackage{graphicx}
\usepackage{subfig}
\usepackage{booktabs}
\usepackage{multirow}
\usepackage{bm}
\usepackage{url}

\usepackage{amsmath}
\usepackage{amssymb}
\usepackage{mathtools}
\usepackage{amsthm}
\usepackage{tikz}
\theoremstyle{plain}
\newtheorem{theorem}{Theorem}[section]

\theoremstyle{definition}
\newtheorem{definition}[theorem]{Definition}

\theoremstyle{remark}

\DeclareMathOperator*{\argmin}{arg\,min}
\DeclareMathOperator*{\E}{\mathbb{E}}

\newcommand{\rarrow}[0]{%
\parbox{0.15cm}{\tikz{\draw[->](0,0)--(0.15cm,0);}}
}
\newcommand{\larrow}[0]{%
\parbox{0.15cm}{\tikz{\draw[<-](0,0)--(0.15cm,0);}}
}

\newcommand{\executeiffilenewer}[3]{%
\ifnum\pdfstrcmp{\pdffilemoddate{#1}}%
{\pdffilemoddate{#2}}>0%
{\immediate\write18{#3}}\fi%
}

\begin{document}

\title{Learning Representative Trajectories of Dynamical Systems via Domain-Adaptive Imitation}

\author{\name Edgardo Solano-Carrillo \email        edgardo.solanocarrillo@dlr.de \\
\name Jannis Stoppe \email jannis.stoppe@dlr.de \\
       \addr German Aerospace Center (DLR) \\ 
       Institute for the Protection of Maritime Infrastructures\\
       Fischkai 1, 27572 Bremerhaven - Germany}


\maketitle

\begin{abstract}
Domain-adaptive trajectory imitation is a skill that some predators learn for survival, by mapping dynamic information from one domain (their speed and steering direction) to a different domain (current position of the moving prey). An intelligent agent with this skill could be exploited for a diversity of tasks, including the recognition of abnormal motion in traffic once it has learned to imitate representative trajectories. Towards this direction, we propose DATI, a deep reinforcement learning agent designed for domain-adaptive trajectory imitation using a cycle-consistent generative adversarial method. Our experiments on a variety of synthetic families of reference trajectories show that DATI outperforms baseline methods for imitation learning and optimal control in this setting, keeping the same per-task hyperparameters. Its generalization to a real-world scenario is shown through the discovery of abnormal motion patterns in maritime traffic, opening the door for the use of deep reinforcement learning methods for spatially-unconstrained trajectory data mining. 
\end{abstract}

\section{Introduction}
This paper is generally concerned with the problem of learning the distribution of trajectories generated by a dynamical system, when only partial information about its evolution rule is known. Such systems --- evolving as $s_{t+1}=f(s_t)$, with $f:\mathbb{R}^d\rightarrow\mathbb{R}^d$ being an unknown \emph{stochastic} function and $s_0$ having a known distribution --- are obiquitous in science and engineering; a reason why advances in their understanding (which are independent of their state representation) have the potential of impacting a number of research fields. Such a global understanding is one of the goals of this work, which we exemplify by developing a model for learning statistics of $f$ that is benchmarked for a diversity of synthetic systems and then used (with slight modifications) in a real-world scenario with a completely different state representation and geometry.

The starting point for our analysis is recognizing that the state of the system $s_t\in\mathbb{R}^d$ and the representation $\hat{a}_t\in\mathbb{R}^m$ of the partial knowledge of its evolution rule belong to two different geometric manifolds, $\mathcal{S}$ and $\mathcal{A}$, respectively. These are connected by a known \emph{deterministic} function $g: \mathcal{S}\times \mathcal{A}\rightarrow\mathcal{S}$ defining such a knowledge, in a way that makes $\hat{s}_{t+1}=g(\hat{s}_t, \hat{a}_t)$ approximate the state $s_{t+1}$ at each time step. In our formulation, the stochasticity of $f$ is taken on by the random variables $\hat{a}_t$ and, since $g$ is not necessarily bilinear, a wide range of systems may be considered. By observing an ensemble of sequences $s_{0:t}$ of states, our interest is then to learn the distribution of the corresponding sequences of \emph{decision} variables $\hat{a}_{0:t}$ that generate the evolution of such states.

An example of the meaning of such formulation is provided by a cheetah learning to chase gazelles. In a given trial, the cheetah has to decide --- from a snapshot of the current position $s_t$ of the gazelle --- the velocity vector $\hat{a}_t$ making its position $\hat{s}_t$ follow $s_t$ as close as possible. Certainly, after many trials, the cheetah learns how to infer $\hat{a}_t$ from complex environmental cues, after crucially discovering simple kinematics laws encapsulated by $g$. We refer to this learning process as \emph{domain-adaptive trajectory imitation}, since it involves learning from a distribution of trajectories $s_{0:t}$ of a dynamical system, by adapting information from one domain $\mathcal{A}$ to a different one $\mathcal{S}$, using partial knowledge of the evolution of $s_t$ encoded by $g$. Since this task engages a decision maker, our approach is suitable for deep reinforcement learning methods \shortcite{Lazaridis2020DeepRL}.

Although the ideas developed here may be extended to any dynamical system, we are motivated by a concrete practical application: \emph{the detection of anomalies from real-time tracking data}. In particular, the results in this paper lead to a source of information for anomaly detection in maritime traffic that is alternative to the one that we have already exploited from a computer vision perspective \shortcite{solano2021}, and then of great potential for maritime situational awareness. It may be useful in the detection of motion patterns corresponding to illegal activities, close in spirit with using agent-based simulations to match the empirical spatio-temporal distribution of crime locations from large-scale human activity data \shortcite{Ross2020SimulatingOM}. Our approach is principled: detecting abnormal behavior by first \emph{learning the distribution} of what is considered normal behavior and then measuring deviations from this at inference time. For this reason, it may be applied to time series featuring different caracteristics (e.g., non-stationarity, irregular sampling rate, missing points, etc.) for which a plethora of different methods have been proposed per characteristic \shortcite{tseries2021}. 

Our key contributions in this paper are therefore:

\begin{itemize}
 \item We introduce an OpenAI Gym environment  \shortcite{brockman2016openai} supporting arbitrary families of reference trajectories (i.e. solutions to the mechanics of abstract dynamical systems) for the trajectory imitation problem. Four built-in families are provided for benchmarking existing and new learning methods;
 \item We propose a robust method (DATI) for learning representative trajectories of dynamical systems by framing the trajectory imitation task as a \emph{reinforced} style transfer problem from the reference trajectories to the rollouts of the reinforcement learning agent -- inspired by image to image translation \shortcite{cycleGAN};
 \item We explore, for the first time, the application of deep reinforcement learning for spatially-unconstrained trajectory data mining; in particular, anomaly detection from tracking data, using maritime traffic as a testbed.
\end{itemize}

The presentation of this work is structured in such a way as to highlight how a single model can learn the main statistical properties of a variety of dynamical systems: from synthetic to a real-world application, keeping (nearly) the same architecture and hyperparameters. 

\section{Related work}
Since the work on dynamical systems is vastly represented in many research fields, here we restrict only to recent methodologies which inspire our present viewpoint.

\textbf{Motion imitation.} At the core of our approach is the acquisition of locomotive skills by an agent that learns to imitate motion. There has recently been increasing interest on this. \shortciteA{LearnToWalk19} taught a robot how to walk from scratch with minimal per-task hyperparameter tuning and a modest number of trials to learn. A similar task was carried out by a drone learning how to fly to a goal marker \shortcite{learn2fly}. More agile locomotion skills have been learned by imitating (from video motion capture) animals \shortcite{RoboImitationPeng20}, complex human acrobatics \shortcite{2018-TOG-SFV}, basketball dribbling \shortcite{basketball}; and by 
simulating realistic human motion from a model of the muscle contraction dynamics \shortcite{sim_human_muscles2019}. The above methods use agents with a fixed embodiment. \shortciteA{core_skills} have proposed a learning framework for core locomotive skills that work for wide variety of legged robots keeping hyperparameter setting and reward scheme. More aligned with our domain-adaptive approach is learning from experts which are different from the agents due to a mismatch of viewpoint, morphology or dynamics \shortcite{dripta2021,dail2020}. Nevertheless, none of these methods deal with the problem of imitating center-of-mass trajectories with significant spatiotemporal extension and complex shapes, as we pursue in this paper. 

\textbf{Trajectory control.} The problem of making a vehicle follow a pre-defined path in space is mainly approached in two ways: trajectory tracking \shortcite{Lee2017}, which demands tracking a timed reference signal; and path following \shortcite{Rubi2020}, where the time dependency is removed and only the geometry is considered. Applications are mostly found in the control of multirotor unmanned aerial vehicles; although moving object grasping is of interest to realize intelligent industrial assembly lines \shortcite{CHEN202162}. Using learning-based methods, these applications include adapting the popular DDPG reinforcement learning method \shortcite{ddpg} for solving the path following problem in a quadrotor with adaptive velocity and considering obstacle avoidance \shortcite{Rubi2021,Rubi2021b}. On the other hand, several inverse reinforcement learning approaches have been designed for tracking control \shortcite{Xue2021,Xue2021b}. Our work is closer in spirit to \shortciteR{Choi2017}, where a representative trajectory is extracted from demonstrations and a reward function learned to imitate such trajectory. We train an agent to directly generate the representative trajectories (not necessarily observed in the training set) without learning a reward function though.

\textbf{Trajectory data mining.} The aim is to automatically discover interesting knowledge from trajectory datasets, which are typically generated from social, traffic, and operational dynamics \shortcite{Wang2020mining}. Traditional clustering and classification methods have served as the basis for more in depth pattern mining and anomaly detection. For pattern mining, different methods have been used for different kinds of patterns, such as periodic \shortcite{Li2010,Li2012},  frequent \shortcite{Giannotti06,Giannotti07}, and collective patterns \shortcite{ZhengYu2015}. For anomaly detection, the techniques are often based on clustering methods and its extensions \shortcite{Belhadi2020}; although supervised learning approaches have also been considered \shortcite{Meng2019}. The use of reinforcement learning for anomaly detection has mainly focused on road traffic \shortcite{Min2018}, for which the motion is constrained by the road networks. In a spatially-unconstrained context, such as the maritime traffic, the trajectory shape complexity increases, encouraging the use of computer vision for trajectory classification \shortcite{VisionTrackClassify}, or graph methods for the detection of representative trajectories \shortcite{graphAIS}. We contribute with a novel deep reinforcement learning method capable of discovering both periodic and non-periodic patterns in spatially-unconstrained trajectory data using a single model.

We emphasize here that, although our work is aligned in objectives with the solution of the problem of motion imitation and trajectory control, we are not interested on the design of controllers allowing a safe navigation of an embodied agent (e.g. autonomous vehicle) under uncertain environments. This requires a model of the inertial properties of the vehicles and their interactions with the environments. We rather aim at a disembodied agent which learns to imitate the typical patterns in traffic to then be able to tell the atypical ones.

\section{Background}
In the following, we formulate the learning problem, describe how it can be reframed within the traditional imitation learning and optimal control settings, and introduce DATI in the next section as a novel method to solve the problem.

\subsection{Preliminary}\label{sec:preliminary}
The evolution $s_{t+1}=f(s_t)$ of complex dynamical systems from an initial state $s_0$ is often hard to estimate. The difficulty lies in: 1) our ignorance of their intricate underlying mechanics, and 2) our ignorance of the nature of the noise source. We are interested in learning the joint distribution, $p$, of trajectories $s_{0:t}\equiv(s_0,s_1,\cdots, s_t)$ of a dynamical system, provided we have knowledge of a local approximation to the underlying evolution rule. This is expressed by a deterministic decomposition of the update rule, $\hat{s}_{t+1}=g(\hat{s}_t, \hat{a}_t)$, in terms of random decision variables $\hat{a}_t$. Restricting our focus to Markovian processes, the joint distribution (assumed absolutely continuous and then admitting density\footnote{\label{footnote:grid} We implicity assume the support of all distributions to be divided into a grid with small cell size implied in practice by $\varepsilon$ in definition \ref{def:rtrack}. Notations such as $p_0(s_0)$ then refer to the probability that the initial state is in a cell containing $s_0$.}) can be written as
\begin{equation}\label{eq:joint_p}
 p(s_{0:T})=p_0(s_0)\,\prod_{t=0}^{T-1}\, p_t(s_{t+1}|s_t),
\end{equation}
and the main task is learning how to sample from the \emph{unknown} $p_t(s_{t+1}|s_t)$. Since knowledge of $s_t$ and the true action $a_t$ imply knowledge of $s_{t+1}$ with complete certainty (by means of $g$), the task is equivalent to estimating $p_t(a_t|s_t)$.   
This is pursued by optimizing a policy (neural network) $ \pi_{\theta}^{\rarrow}(\hat{s}_t)\rightarrow \hat{a}_t$ which samples actions from the distribution $p_{\theta}(\hat{a}_t|\hat{s}_t,t)$, and from which the next state is approximated as $\hat{s}_{t+1}=g(\hat{s}_t, \pi_{\theta}^{\rarrow}(\hat{s}_t))\equiv \pi_{\theta}(\hat{s}_t)$. This optimization brings $p_{\theta}(\hat{a}_t|\hat{s}_t,t)$ close to $p_t(a_t|s_t)$ with respect to some distance/divergence discussed later.

\subsection{Problem formulation}
Given an ensemble of trajectories $s_{0:T}\sim p$ generated by a dynamical system, the task is to find a policy $\pi_{\theta}$ that replicates the shape of a \emph{typical} trajectory $s_{0:T}^{*}$ after rolling out $\pi_{\theta}(\hat{s}_{t})\rightarrow \hat{s}_{t+1}$. That is, consider a partition of the time interval $[0,T]$ into $\tau$ equally-spaced timesteps.\footnote{This assumption may be relaxed for real-world applications with irregular sampling rate of the data and varying horizons $T$, as in section \ref{sec:RealWorld}.} Then, the predicted sequence $\hat{s}_{0:T}^\theta\equiv(\hat{s}_{0}, \pi_{\theta}(\hat{s}_{0}),\cdots,\pi_{\theta}(\hat{s}_{T-1}))$ and the reference sequence $s_{0:T}^{*}\equiv(s_{0}^*, s_{1}^*,\cdots,s_{T}^*)$, having the same starting state $\hat{s}_{0}=s_{0}^*$, should match their shapes at optimal $\hat{\theta}$. For concreteness, the shape similarity is with respect to the dynamic time warping distance \shortcite{fastDTW} --- popular for comparing two time series not necessarily aligned in time --- so the optimal model has
\begin{equation}\label{eq:minDTW}
 \hat{\theta} = \argmin_{\theta} D_{\textrm{dtw}}(\hat{s}_{0:T}^\theta, s_{0:T}^{*}).
\end{equation}
In practice, $n$ reference trajectories $s_{0:T}^{*}$ not seen during training of $\pi_{\theta}$ are sampled from $p$ at inference time, and policies trained with $n$ different seeds are rolled out from the corresponding initial conditions. This leads to our notion of typicality, and hence of a learned representative trajectory:

\begin{definition}[Representative trajectory] \label{def:rtrack}
 Given a small $\varepsilon>0$, a representative trajectory $\hat{s}_{0:T}^\theta$ is said to be learned by a policy $\pi_{\theta}$ if there is a reference trajectory $s_{0:T}^*$ (out of the $n$ in the test set) for which $D_{\textrm{dtw}}(\hat{s}_{0:T}^\theta, s_{0:T}^*)<\varepsilon$.
\end{definition}

\noindent Since we will be discussing different methods to bring $\hat{p}_{t}(\hat{a}_t|\hat{s}_t)$ close to $p_t(a_t|s_t)$, as mentioned is \ref{sec:preliminary}, our optimization objective in \eqref{eq:minDTW} is chosen as a common metric to compare these methods. In the following, we focus on a simple 2D representation of the state $s_t$ ($d=2$), as a proof of concept. This allows us to design an intuitive scenario similar to the cheetah chasing a gazelle but which does not end after predation.

\subsection{Mouse and hidden cheese game}
 As a prototype for our synthetic experiments in section \ref{sec:SynExp}, consider a mouse (agent) chasing intermittently-hidden cheese (trajectory demonstrator). At timestep $t$, the mouse has to decide and act $\hat{a}_{t}=(u_{t}, \xi_{t})$ to make its speed $u_{t}$ and steering direction $\xi_{t}$ to take it from its current position $\hat{s}_{t}$ to the currently \emph{unknown} position of the cheese. The latter is only revealed at timestep $t+1$ to be $s_{t+1}\equiv(x_{t+1}, y_{t+1})$, and the mouse is perfectly rewarded if 
\begin{equation}\label{eq:perfect_pi}
 s_{t+1} = g(\hat{s}_{t}, \hat{a}_{t})\equiv \hat{s}_{t}+(u_{t}\cos(\xi_{t}),u_{t}\sin(\xi_{t}))\,dt,
\end{equation}
making it is able to taste the cheese (here $dt=T/\tau$). The second equality in \eqref{eq:perfect_pi} defines $\hat{s}_{t+1}$, so the cheese is tasted when $\hat{s}_{t+1}=s_{t+1}$. Since $\hat{s}_t$ is stochastically generated, it is unlikely that this takes place, unless we impose a minimal distance between $\hat{s}_t$ and $s_t$ (e.g., being within the same cell, as mentioned in footnote \ref{footnote:grid}) within which tasting really happens. In practice, this is implied by the selection of $\varepsilon$ in definition \ref{def:rtrack}.

The function $g$ in this case represents an intuition about kinematics that the mouse has in advance. It is based only on partial information about the true evolution rule prescribed by $s_{t+1}=f(s_t)$, namely, an $O(dt)$ approximation, with a two-dimensional decision space $\mathcal{A}$. A different decomposition is obtained with an $O(dt^2)$ approximation, introducing accelerations as part of the decision variables $\hat{a}_t$. Note that a different geometry of the state space may also be considered, as in Eq. \eqref{eq:onearth}, or even problems not defining physical motion.

The degree to which the decision variables $\hat{a}_t$ (implied by a given decomposition) lead to a satisfactory replication of the shape of reference trajectories leads to the concept of a perfect decision maker:

\begin{definition}[Perfect policy]\label{def:perfect_pol}
 Given a reference trajectory $s_{0:T}^{*}$, a perfect policy rollout $\pi^{*}(\hat{s}_t)\rightarrow\hat{s}_{t+1}$ replicates this trajectory indentically, by means of Eq. \eqref{eq:perfect_pi}. That is, $\hat{s}_t=s_t^{*}$, $\forall t$, making $D_{\textrm{dtw}}(\hat{s}_{0:t}^\theta, s_{0:t}^{*})=0$. This means that the agent has full knowledge of the underlying mechanics of the process, guessing the next state, $s_{t+1}$, and taking corresponding actions $a_t=(u_t,\xi_t)$ with speed $u_t=\|s_{t+1}-s_{t}\|/dt$ and steering angle $\xi_t=\tan^{-1}[(s_{t+1}-s_t)\cdot \hat{y}\,/\,(s_{t+1}-s_t)\cdot \hat{x}]$ in order to get there. 
\end{definition}
Since life is not perfect, the ever hungry and intelligent mouse will learn, after many trials (containing similarly shaped $s_{0:T}$), to discover the main features of the cheese trajectories. At inference, the mouse is fooled with no cheese signal, but its trajectory is collected and compared with the reference $s_{0:T}^*$ to judge how representative it is. In the following, we describe different approaches that we will compare later for the solution of this problem.

\subsection{Imitation learning}
 Imitation learning aims to learn a policy $\pi_{\theta}^{il}(s_t)\rightarrow \hat{a}_t$ mimicking demonstrations $\mathcal{D}=((s_0, a_0), (s_1, a_1), \cdots, (s_T, a_T))$ from an expert whose actions $a_t \sim \pi^{*}$ are collected after observation of many instances of environmental state sequences $s_{0:T}$. Several approaches have been developed for this, see e.g. \shortciteA{survey2021} for a survey. The simplest baseline, which we adopt here for benchmarking the synthetic experiments, is \textit{Behavioral Cloning} (BC). This learns the policy by considering the setting as a supervised regression problem over $\mathcal{D}$ \shortcite{Pomerleau91,pmlr-v9-ross10a}. That is, the unknown transition distribution $p_t(a_t|s_t)$ of relevance for \eqref{eq:joint_p} is estimated as $p_{\theta}(\hat{a}_t|s_t)$ after minimizing the Kullback-Leibler divergence between the two. This amounts to a maximum likelihood optimization of $p_{\theta}(\hat{a}_t|s_t)$.

The application of these method to the present problem demands the interpretation of the environment states $s_t$ as comprising the trajectories $s_{0:T}$ to be imitated. The actions $a_t$ in the demonstration set $\mathcal{D}$ are generated by the best expert: a perfect policy $\pi^{*}$, according to definition \eqref{def:perfect_pol}. At inference, when the observations $s_{0:t}$ are removed, the predicted $\hat{s}_{t}$'s are used instead when rolling out the learned policy, i.e. $\hat{s}_{t+1}=g(\hat{s}_t, \pi_{\theta}^{il}(\hat{s}_t))=\pi_{\theta}(\hat{s}_t)$. 

\subsection{Reinforcement learning using sparse rewards}\label{sec:DDPG} 
In the standard reinforcement learning framework, the agent-environment interaction is modeled as a Markov decision process $(\mathcal{S}, \mathcal{A}, \mathcal{P}, r, \gamma, p_0)$, where $\mathcal{S}$ and $\mathcal{A}$ are the state and action spaces, respectively, $\mathcal{P}(s'|a,s)$ is the transition distribution of the environment, $r=r(s,a)$ is the reward function, $\gamma\in(0,1)$ is the discount factor, and $p_0=p_0(s)$ is the initial state distribution of the environment.

As a motivation behind our method, we adapt the DDPG agent \shortcite{ddpg} to the trajectory imitation problem (denoted as DDPG-TI). This still consists of two actor-critic models --- the learned $(\pi_{\theta}^l, c_{\theta_c}^l)$ and target $(\pi_{\beta}^r, c_{\beta_c}^r)$ networks, the latter slowly tracking the former during training --- the actors guided by the return $R_t=\sum_{i=t}^{T}\gamma^{i-t}\,r(s_i, a_i)$ from a state. However, instead of having tuples $(s_t,a_t,r_t,s_{t+1})$ in the replay buffer during maximization of the expected return from the starting distribution, we use $(\hat{s}_t,\hat{a}_t,r_t,\hat{s}_{t+1},t)$, where $t$ is processed by a scale-invariant embedding of time \shortcite{t2v} which is shared by the networks. The reward signal here is $r_t=\pm1$ according to whether or not $\tilde{D}_{\textrm{dtw}}(\hat{s}_{0:t}^\theta, s_{0:t})<\varepsilon$, where the tilde denotes an exponentially-smoothed and normalized dynamic time warping distance. This is inspired by the constant rewards used by \shortciteA{Reddy2020}.

In this formulation, the agent does not make decisions $\hat{a}_t$ during training based on the current state $s_t$, but rather on its current prediction of that state $\hat{s}_t$. This change is necessary to avoid the DDPG-TI agent taking the same actions regardless of the current state, an observation from early experimentation which motivated the introduction of the embedding of time. At inference, the learned policy leads to the predictions $\hat{s}_{t+1}=g(\hat{s}_t, \pi_{\theta}^{l}(\hat{s}_t, t)+\eta_t)=\pi_{\theta}(\hat{s}_t)$, where $\eta_t$ is the noise used for exploration of the environment (originally taken as an Ornstein-Uhlenbeck process by \shortciteR{ddpg}).

\section{Domain-adaptive trajectory imitation}
Since we are interested in learning the distribution of the trajectories to be imitated, adding the noise $\eta_t$ to the output of the actor entails a fake stochasticity in the results. So, building from DDPG-TI, the first observation is to change the actor  $\pi_{\theta}^{l}(\hat{s}_t,t)\rightarrow \hat{a}_t+\eta_t$ to the actor $\pi_{\theta}^{\rarrow}(\hat{s}_t, \eta_t, t)\rightarrow \hat{a}_t$, taking us to the realm of generative models; in particular, the adversarial generative models (GANs) which recover the data distribution \shortcite{Goodfellow17}.  

With the above observation in mind, the main idea behind DATI may be informally stated as considering the imitation of $s_{0:t}$ by $\hat{s}_{0:t}$ as a \emph{style} transfer problem between the domains spanned by these set of trajectories. Cycle-consistent generative adversarial networks (CycleGAN) \shortcite{cycleGAN} have shown significant success in the style transfer task for unpaired image-to-image translation, so we extend them here to the domain-adaptive trajectory imitation problem, adding a novel reinforcement signal.

As with DDPG-TI, we have two actor-critic networks: $(\pi_{\theta}^{\rarrow}, c_{\theta_c}^{\rarrow})$ and $(\pi_{\beta}^{\larrow}, c_{\beta_c}^{\larrow})$. However, each pair is now trained adversarially using the Wasserstein loss \shortcite{wassersteinGAN}.  Specifically, for the first pair, we take an actor $\pi_{\theta}^{\rarrow}(\hat{s}_t, \eta_t, t)\rightarrow \hat{a}_t$ which learns how to sample from the distribution $p_{\theta}(\hat{a}_t|\hat{s}_t, t)$ (sometimes just denoted $\hat{p}_t(\hat{a}_t|\hat{s}_t)$) by accesing the noise prior $p_{\eta_t}$. This is accomplished by valuing the chosen decisions with a critic $c_{\theta_c}^{\rarrow}(\hat{a}_t, r_t, t)$ --- lying in the space of 1-Lipschitz functions\footnote{In our  experiments, this condition is kept by adopting the method from \shortciteA{improved_wgan}.} (denoted $\|c_{\theta_c}^{\rarrow}\|_L \le 1$), and $r_t$ being the sparse rewards of section \ref{sec:DDPG} --- and both networks optimized as
\begin{equation}\label{eq:wgan}
\begin{split}
 \max_{\|c_{\theta_c}^{\rarrow}\|_L \le 1}\; & \E_{a_t\sim p_t(a_t|s_t)}[c_{\theta_c}^{\rarrow}(a_t)]-\E_{\eta_t\sim p_{\eta_t}}[c_{\theta_c}^{\rarrow}(\pi_{\theta}^{\rarrow}(\eta_t))],\\
 & \hspace{0.5cm}\ \max_{\pi_{\theta}^{\rarrow}}\; \E_{\eta_t\sim p_{\eta_t}}c_{\theta_c}^{\rarrow}(\pi_{\theta}^{\rarrow}(\eta_t)),
 \end{split}
\end{equation}
where we have omitted some arguments of the functions for simplicity. That is, the actor is trained to maximize the value that the critic assigns to its decisions (bottom of \eqref{eq:wgan}), whereas the critic is trained to separate this value from the value of perfect decisions (top of \eqref{eq:wgan}). These perfect decisions, $a_t$, are those from a perfect policy $\pi^{*}$ according to definition \ref{def:perfect_pol}.

This optimization procedure guarantees --- under plausible continuity assumptions for the actor --- that the distribution $p_{\theta}(\hat{a}_t|\hat{s}_t, t)$ converges to the true distribution $p_t(a_t|s_t)$ with respect to the Wasserstein (a.k.a Earth Mover) distance \shortcite{wassersteinGAN}. As mentioned, knowing the true $a_t$ reproduces the next state as $s_{t+1}=g(s_t, a_t)$, so the $p_t(a_t|s_t)$ estimated by DATI through $p_{\theta}(\hat{a}_t|\hat{s}_t, t)$ basically leads to an approximation of the $p_t(s_{t+1}|s_t)$ that are of interest in \eqref{eq:joint_p}. 

The networks in the second pair, $(\pi_{\beta}^{\larrow}, c_{\beta_c}^{\larrow})$, have the same architecture as their respective networks in the first pair. However, the actor $\pi_{\beta}^{\larrow}(\hat{a}_t, \eta_t,t)\rightarrow \hat{s}_t^{\larrow}$ is concurrently trained to \emph{undo} the action of $\pi_{\theta}^{\rarrow}(\hat{s}_t, \eta_t, t)\rightarrow \hat{a}_t$, i.e. by reconstructing the $\hat{s}_t$ that is targeted by $\hat{s}_t^{\larrow}$ . This ``backward'' actor, $\pi_{\beta}^{\larrow}$, is valued by a critic $c_{\beta_c}^{\larrow}(\hat{s}_t^{\larrow}, r_t, t)$, both trained similar to \eqref{eq:wgan}

\begin{equation}
\begin{split}
 \max_{\|c_{\beta_c}^{\larrow}\|_L \le 1}\; & \E_{\hat{s}_t\sim \hat{p}_t(\hat{s}_t|\hat{a}_{t})}[c_{\beta_c}^{\larrow}(\hat{s}_t)]-\E_{\eta_t\sim p_{\eta_t}}[c_{\beta_c}^{\larrow}(\pi_{\beta}^{\larrow}(\eta_t))],\\
 & \hspace{0.5cm}\ \max_{\pi_{\beta}^{\larrow}}\; \E_{\eta_t\sim p_{\eta_t}}c_{\beta_c}^{\larrow}(\pi_{\beta}^{\larrow}(\eta_t)),
 \end{split}
\end{equation}
where samples from $\hat{p}_t(\hat{s}_t|\hat{a}_{t})$ are obtained by rolling out the ``forward'' policy $\pi_{\theta}^{\rarrow}$. This requires further explanation, since we we have said in the previous paragraph that $\pi_{\theta}^{\rarrow}$ samples from the distribution $\hat{p}_t(\hat{a}_t|\hat{s}_t)$, not the posterior $\hat{p}_t(\hat{s}_t|\hat{a}_{t})$. The question is: if $\pi_{\theta}^{\rarrow}(\cdot, \eta_t, t)$ maps $\hat{s}_t$ deterministically into $\hat{a}_t$, is the value of $\hat{s}_t$ implied by having only knowledge of $\hat{a}_t$ (and, of course, of $\eta_t, t$)? The answer is yes, provided that $\pi_{\theta}^{\rarrow}(\cdot, \eta_t, t)$ is a \emph{bijective} mapping. Therefore, whenever a value of $\hat{a}_t$ is fetched from the replay buffer --- which collects the tuples $(\hat{s}_t,\hat{a}_t, a_t, r_t,\eta_t,t)$ --- there is only one possible value of $\hat{s}_t$ which produced that value of $\hat{a}_t$ using $\pi_{\theta}^{\rarrow}$, namely, the one recorded in the same tuple.

The bijective nature of the actors $\pi_{\theta}^{\rarrow}$ and $\pi_{\beta}^{\larrow}$ is enforced using \emph{cycle consistency}. That is, by making both $\pi_{\beta}^{\larrow} \circ \pi_{\theta}^{\rarrow}$ and $\pi_{\theta}^{\rarrow}\circ \pi_{\beta}^{\larrow}$ approximate the identity mapping: 

\begin{equation}\label{eq:cycle}
\begin{split}
\min_{\pi_{\theta}^{\rarrow}, \pi_{\beta}^{\larrow}}\; & \E_{\hat{a}_t\sim \hat{p}_t(\hat{a}_t|\hat{s}_{t})} \| \pi_{\theta}^{\rarrow}(\pi_{\beta}^{\larrow}(\hat{a}_t)) - \hat{a}_t\|_1 ,\\
 \min_{\pi_{\theta}^{\rarrow}, \pi_{\beta}^{\larrow}}\; & \E_{\hat{s}_t\sim \hat{p}_t(\hat{s}_t|\hat{a}_{t})} \| \pi_{\beta}^{\larrow}(\pi_{\theta}^{\rarrow}(\hat{s}_t)) - \hat{s}_t\|_1.
 \end{split}
\end{equation}

Note that the bottom of \eqref{eq:cycle} is a statement of minimization of the reconstruction error $\|\hat{s}_t^{\larrow}- \hat{s}_t\|_1$: the output of $\pi_{\beta}^{\larrow}$ is supervised by confronting it against the ground truth $\hat{s}_t$. To achieve a comparable supervision for $\pi_{\theta}^{\rarrow}$, we enforce its output $\hat{a}_t$ to approximate the ground truth $a_t$ via $L_1$ penalty:

\begin{equation}
\min_{\pi_{\theta}^{\rarrow}}\; \E_{\hat{s}_t\sim \hat{p}_t(\hat{s}_t|\hat{a}_{t}), \;a_t\sim p_t(a_t|s_t)} \| \pi_{\theta}^{\rarrow}(\hat{s}_t) - a_t\|_1,
\end{equation}
with the real $a_t$'s being obtained, as before, from a perfect policy $\pi^{*}$ according to definition \ref{def:perfect_pol}. At inference, the ``forward'' actor is used for the trajectory predictions according to $\hat{s}_{t+1}=g(\hat{s}_t, \pi_{\theta}^{\rarrow}(\hat{s}_t,\eta_t,t))=\pi_{\theta}(\hat{s}_t)$ 

Finally, the motivation for using the scale-invariant embedding of time by \shortciteA{t2v} in our method is to obtain a positional encoding of the time series representing the trajectories which is able to capture both periodic and non-periodic patterns in the data, regardless of whether we use the index $i$ of the time $t=i\,dt$ of each event. This will prove to be beneficial for our method from the ablation studies in section \ref{subsec:ablation}.
\begin{figure*}[ht]
 \centering
 \subfloat{{\includegraphics[scale=0.6]{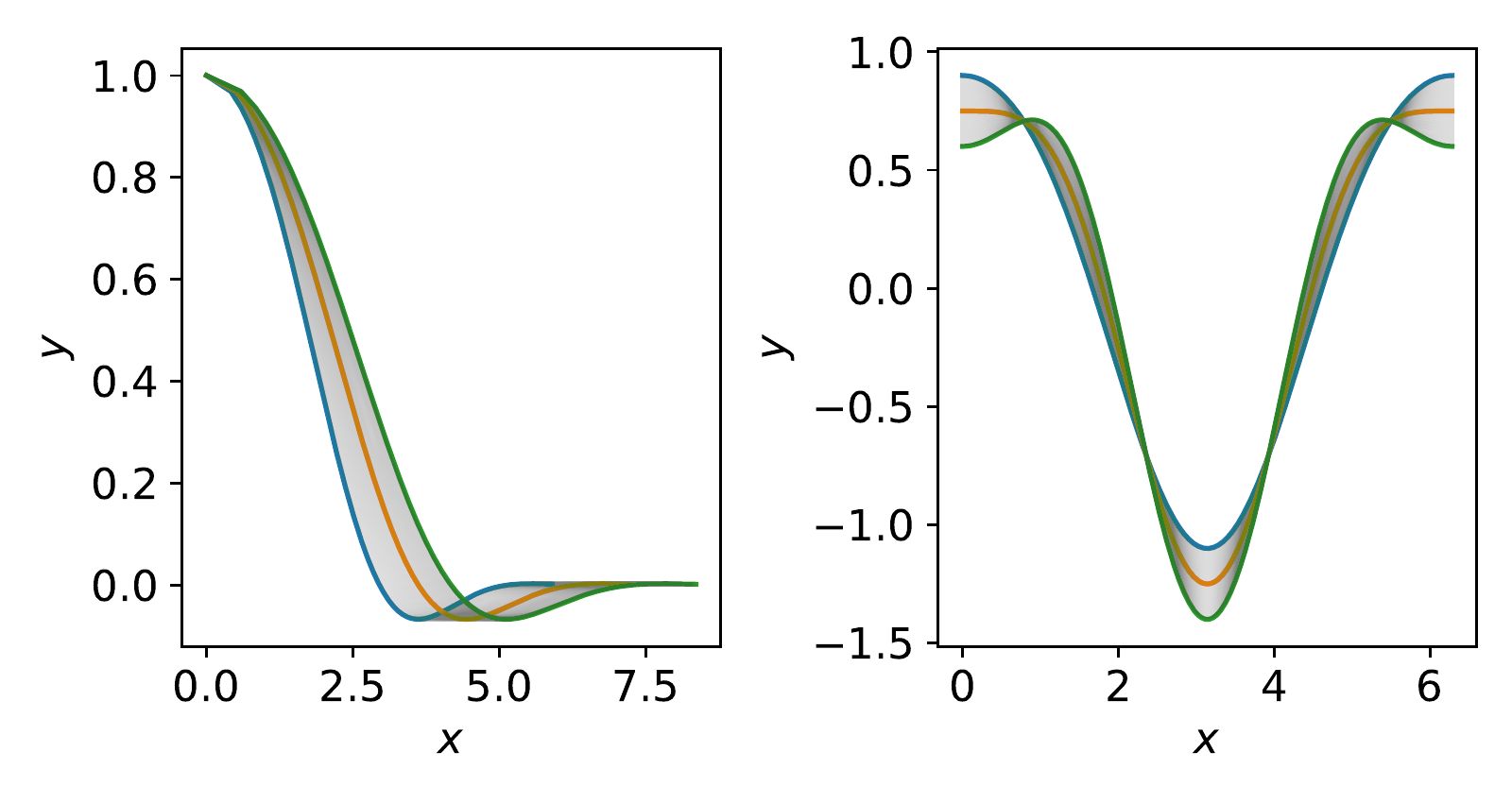}}}\\
 \vspace*{-0.3cm}\ 
 \subfloat{{\includegraphics[scale=0.64]{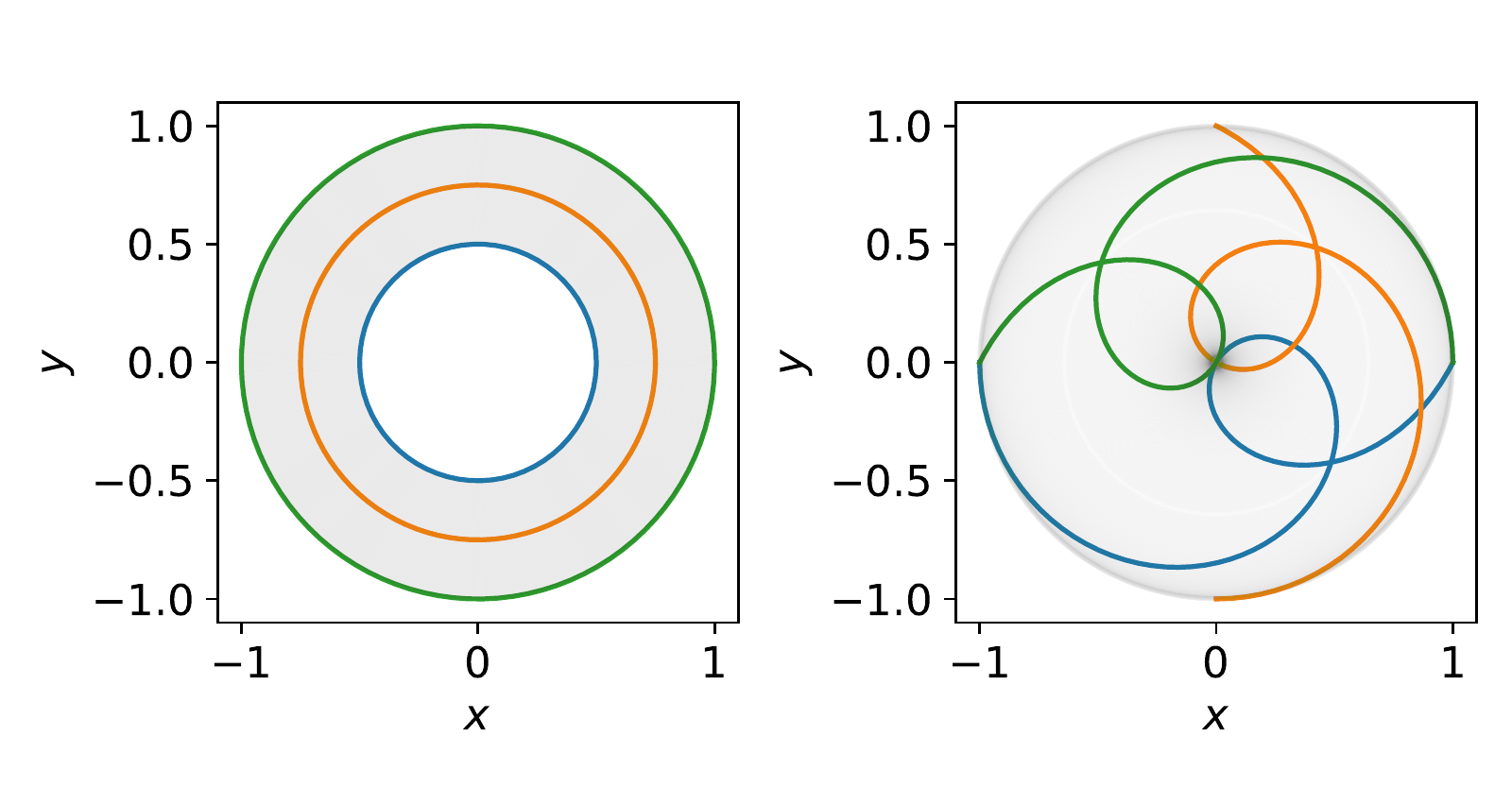}}}
 \caption{Family of trajectories $\mathcal{F}_{\bm{\alpha}}$ considered for the domain-adaptive imitation task (from top left to bottom right: FixedStart, UShaped, Circles and Ribbons). These may be thought of as solutions to abstract dynamical systems with a two-dimensional state representation.}
 \label{fig:family} 
\end{figure*} 

\section{Synthetic experiments}\label{sec:SynExp}
The synthetic experiments are carried out on a variety of families of solutions to the mechanics of hypothetical dinamical systems. These are characterized by the initial state distribution $p_0(s_0)$ in \eqref{eq:joint_p}, and a prescription to generate the subsequent states $s_{1:t}$ of the trajectory deterministically, given $s_0$. The corresponding ensembles are denoted by $\mathcal{F}_{\bm{\alpha}}$: comprising a family of trajectories with shape parameters $\bm{\alpha}=\{\alpha, \cdots\}$ --- all fixed except $\alpha$, which is sampled uniformly within given intervals for each episode.  

We provide an OpenAI Gym  environment \shortcite{brockman2016openai} supporting (arbitrary) user-defined ensembles, $\mathcal{F}_{\bm{\alpha}}$, serving as demonstrations $s_{0:T}$ of the cheese signals for the mouse. We focus on four families defined in table \ref{tab:families} and visualized in Fig. \ref{fig:family}:

1) \emph{FixedStart} provides trajectories with the same starting point, having an inflection point near the beginning of the journey and then possibly confusing gradient descent methods aiming only at learning the mapping $x\rightarrow y$ (geometry without the time component).

2) \emph{UShaped} provides trajectories with a reflection symmetry about a vertical axis passing through their lowest points, and having varying starting and ending features. 

3) \emph{Circles} provide the simplest expression of a periodic pattern in the trajectories. Nevertheless, they are complex enough, since the agents must keep their speed constant. 

4) \emph{Ribbons} provide trajectories to test how the networks disentangle space and time: by presenting a point in space that is visited twice with different headings.

{\def\arraystretch{1.5}\tabcolsep=3.5pt
\begin{table}[t]
\centering
\begin{tabular}{ccc}
\toprule
\multirow{2}{*}{Family $\mathcal{F}_{\bm{\alpha}}$} &
\multicolumn{2}{c}{$s_t= (x_t, y_t)$}\\
 & $x_t$ & $y_t$  \\
\midrule

FixedStart & $\sqrt{\alpha t}$ & $ \cos(\omega t)\,e^{-\kappa t}$\\
UShaped & $\omega t$ & $\cos(\omega t)-\alpha \cos(2\omega t)/2$\\
Circles &$\alpha \cos(\omega t)$ & $\alpha \sin(\omega t)$\\
Ribbons &\begin{tabular}{@{}c@{}}$R_1-R_2 \cos(\omega t/4)$ \\ $\times\cos(\omega t+\alpha)$\end{tabular}& \begin{tabular}{@{}c@{}}$R_1-R_2 \cos(\omega t/4)$ \\ $\times\sin(\omega t+\alpha)$\end{tabular}\\

\bottomrule
\end{tabular}
\caption{Defining equations for the ensembles of trajectories $\mathcal{F}_{\bm{\alpha}}$ used for benchmarking models. For the fixed values of the shape parameters complementary to $\alpha$, see appendix \ref{subsec:shape_param}.  }
\label{tab:families}
\end{table}
} 

\subsection{Experimental setup and results}
\begin{figure*}[ht]
 \centering
  \subfloat{{\includegraphics[scale=0.6]{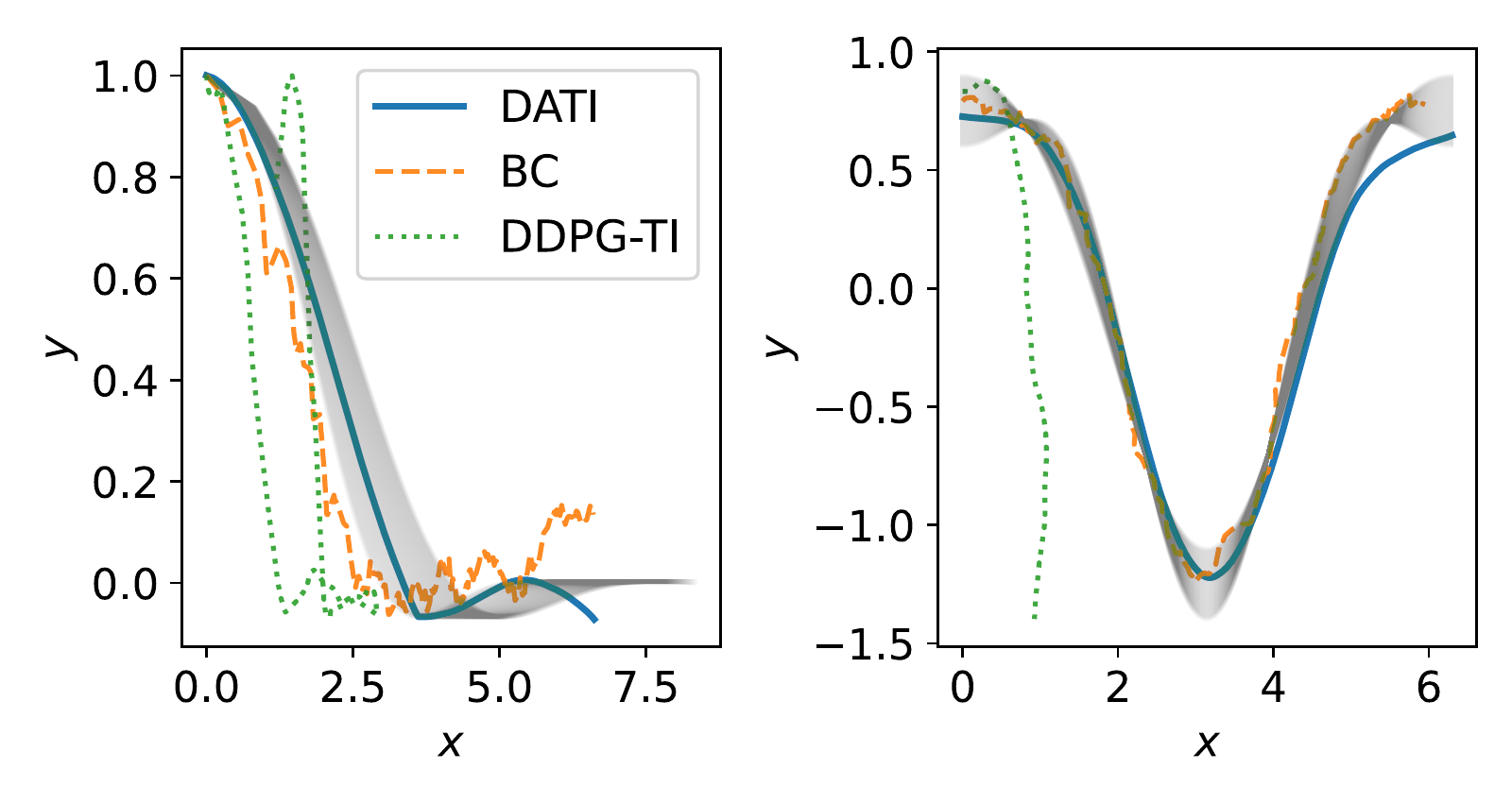}}}\\
 \vspace*{-0.3cm}\ 
 \subfloat{{\includegraphics[scale=0.64]{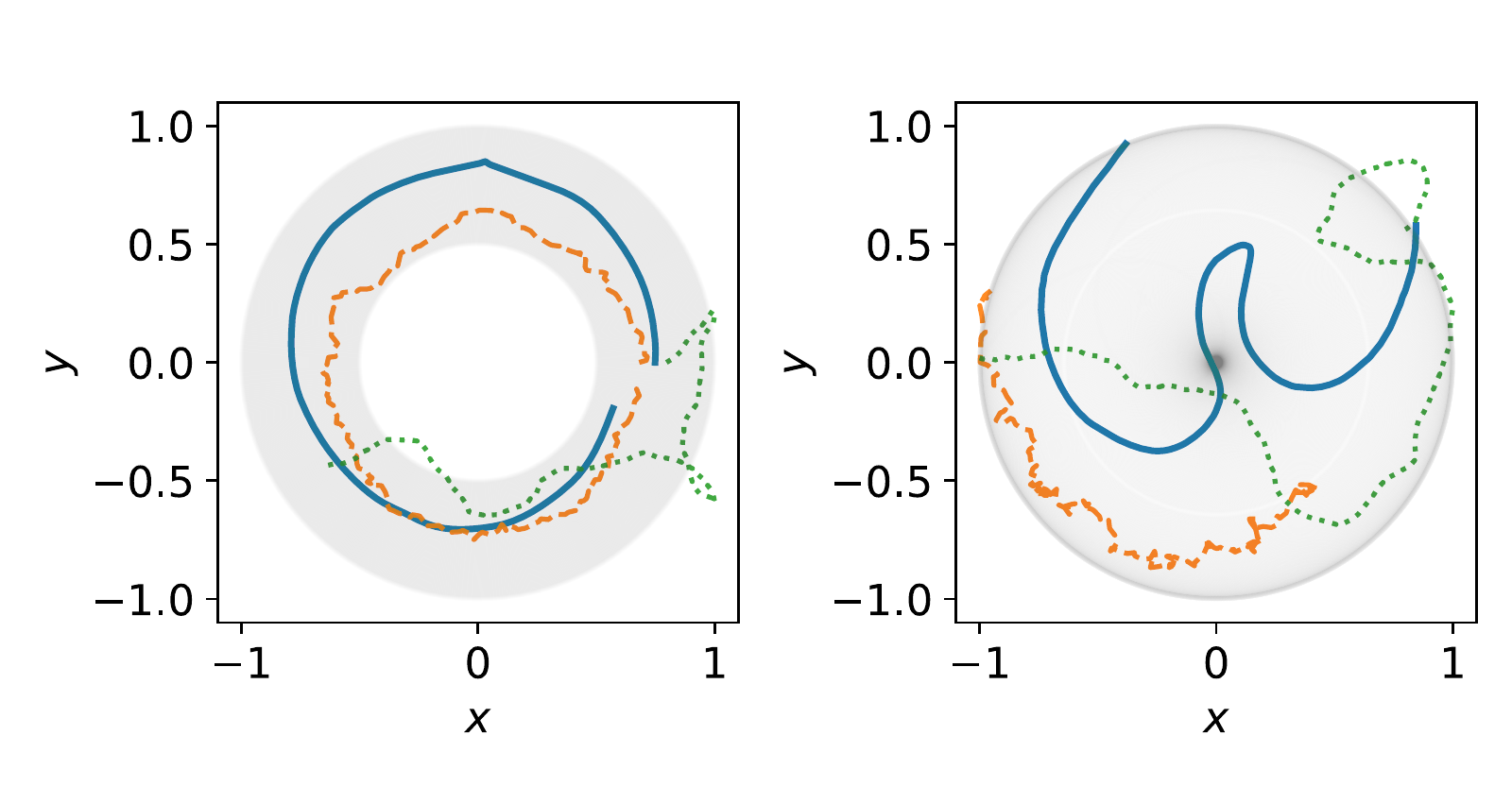}}}

 \caption{Rollouts $\pi_{\theta}(\hat{s}_t)\rightarrow \hat{s}_{t+1}$ of the learned policies for different methods at a seed giving the best dynamic time warping distance $\tilde{D}_{\textrm{dtw}}(\hat{s}_{0:T}^\theta, s_{0:T}^*)$ to a reference trajectory $s_{0:T}^*$ (with $\hat{s}_0=s_0^*$), out of 10 trials. The results correspond, from top left to bottom right, to the families of trajectories $\mathcal{F}_{\bm{\alpha}}$ with FixedStart, UShaped, Circles, and Ribbons of Fig. \ref{fig:family}. For a quantitative assessment, see table \ref{tab:results}.}
 \label{fig:results} 
\end{figure*}

The experiments are carried out targeting robustness of the models to the variation of the learning tasks. So we keep the same hyperparameters for all the families of trajectories. Additionally, we impose a limit in the number of transitions for the models to learn the requested task. That is, we sample 100 episodes (a trajectory from $\mathcal{F}_{\bm{\alpha}}$ per episode), each with a number of $\tau=200$ timesteps. Code is supplementary provided for reproducibility.    

\textbf{Architectures}. The actor-critic models for DATI are multilayer perceptrons with feature extractors having 16 units, time embedding dimension of 76, and 4 hidden layers -- before the network outputs -- having 32 units each. The critic networks process the rewards $r_t\in\{-1,1\}$ by tiling the input to have the dimensionality of the sum of all other extracted-feature dimensions, passing this to a dense layer with non-negative kernel contraints and concatenating the output with such other extracted features before entering the 4 hidden layers previous to the (elu-activated) network output. The novel idea behind is that this propagates through the critic networks the signal of having low  output for negative rewards and high output for positive rewards. Finally, the latent dimension of $\eta_t$ for the actors is 64. 

DDPG-TI has similar actor-critic architectures, but since the actors do not include a latent dimension, an increase of the feature extractors to 80 units is needed for their 4 hidden units before the output to observe the same input dimensionality as DATI. On the other hand, BC is trained using the \texttt{imitation} library \shortcite{wang2020imitation}. In order to get actor architectures comparable to DDPG-TI and DATI, feature extractors with 156 units are used (making up the time embedding $+$ latent dimensions $+$ 16), followed by 4 hidden layers of 32 units. For the maximum likelihood estimation of $p_{\theta}(\hat{a}_t|s_t)$, the latent features before the output of $\pi_{\theta}^{il}$ are linearly transformed to define the mean and diagonal covariance of a Gaussian distribution $p_{\theta}(\hat{a}_t|s_t)$ network. Hyperparameter selection for the different methods may be found in appendix \ref{subsec:hyper-param}.

\textbf{Results}. The models are trained with 10 seeds and evaluated with respect to 10 random reference trajectories $s_t^*$ not seen in the training set. We compare the performance by reporting the best exponentially smoothed and normalized dynamic time warping distance $\tilde{D}_{\textrm{dtw}}(\hat{s}_{0:T}^\theta, s_{0:T}^*)$ over the 10 test trials. The normalization constant is the dynamic time warping ``diameter'' $D_{\textrm{dtw}}^>$, defined as the distance between the blue and green boundary trajectories in Fig. \ref{fig:family}. The results are shown in table \ref{tab:results}, with a visualization of the shapes of the best trajectories attained at inference in Fig. \ref{fig:results}.  

{\tabcolsep=5pt
\begin{table}[hb]
\centering
\begin{tabular}{ccccc}
\toprule
  & FixedStart & UShaped & Circles & Ribbons \\
\midrule

DDPG-TI &0.489 & 2.138& 0.972 & 0.172 \\
BC & 0.124& 0.365& 0.106& 0.113\\
\textbf{DATI} & $\bm{0.065}$ & $\bm{0.231}$& $\bm{0.058}$& $\bm{0.033}$ \\

\bottomrule
\end{tabular}
\caption{Lowest $\tilde{D}_{\textrm{dtw}}(\hat{s}_{0:T}^\theta, s_{0:T}^*)$ over 10 test trials.}
\label{tab:results}
\end{table}
}
 Clearly, DATI is able to generate trajectories that look closer to the ones in the families $\mathcal{F}_{\bm{\alpha}}$. Note that DATI and DDPG-TI were both trained using the reward signal $r_t=\pm1$ according to whether or not $\tilde{D}_{\textrm{dtw}}(\hat{s}_{0:t}^\theta, s_{0:t})<\varepsilon$. For the experiments, the $\varepsilon$ in definition \ref{def:rtrack} is taken as $10\%$ of the dynamic time warping diameter $D_{\textrm{dtw}}^>$. So only DATI is able to learn representative trajectories in most of the cases (numbers below 0.1 in table \ref{tab:results}). We further investigate now the importance of some design choices in the actor-critic models in DATI. 

\subsection{Ablation study}\label{subsec:ablation}
\begin{table}[t]
\centering
\begin{tabular}{ccc}
\toprule
\multirow{2}{*}{Transformation} &
\multicolumn{2}{c}{$\tilde{D}_{\textrm{dtw}}(\hat{s}_{0:T}^\theta, s_{0:T}^*)$} \\
 & top-1 & top-2 \\
\midrule

Original setup & $\bm{0.058}$ & $\bm{0.074}$\\
No time embedding & 1.829 & 1.976\\
No reward reinforcement &0.305 & 0.451\\

\bottomrule
\end{tabular}
\caption{Changes in the metric used in table \ref{tab:results} under ablation of some layers in the actor-critic models in DATI. The experiments are carried out for the family of Circles trajectories.}
\label{tab:ablation}
\end{table}
We take the family of Circles trajectories and apply the following transformations --- all other things being equal --- to the architecture of the actor-critic models in DATI:
\begin{itemize}
 \item \emph{No time embedding}: we remove the notion of time    exogenously imposed on the networks.
 \item \emph{No reward reinforcement}: we do not let the critics know about the goal of minimizing the dynamic time warping distance between actor rollouts and demonstrations.
\end{itemize}

In order to get sensible statistical results \shortcite{agarwal2021deep} of the effect of these changes, it suffices to monitor the best and second best $\tilde{D}_{\textrm{dtw}}(\hat{s}_{0:T}^\theta, s_{0:T}^*)$ over the 10 test trials. These are shown in table \ref{tab:ablation}. We observe a significant negative impact on the performance when the time embedding is removed. Similarly, removing the reward signal has appreciable negative effects. These are therefore essential ingredients in the design of DATI.

The conclusion from these experiments is that DATI is a successful method for learning representative trajectories, being robust to changes in their geometries --- i.e. keeping the same architecture and hyperparameters it can represent a rich set of spatiotemporal phenomena. In order to further test this, we evaluate its generalization to a real world scenario, taking maritime traffic as an example. 
 
\section{Real-world experiments}\label{sec:RealWorld}
\begin{figure}[t]
 \centering
 \includegraphics[scale=0.25]{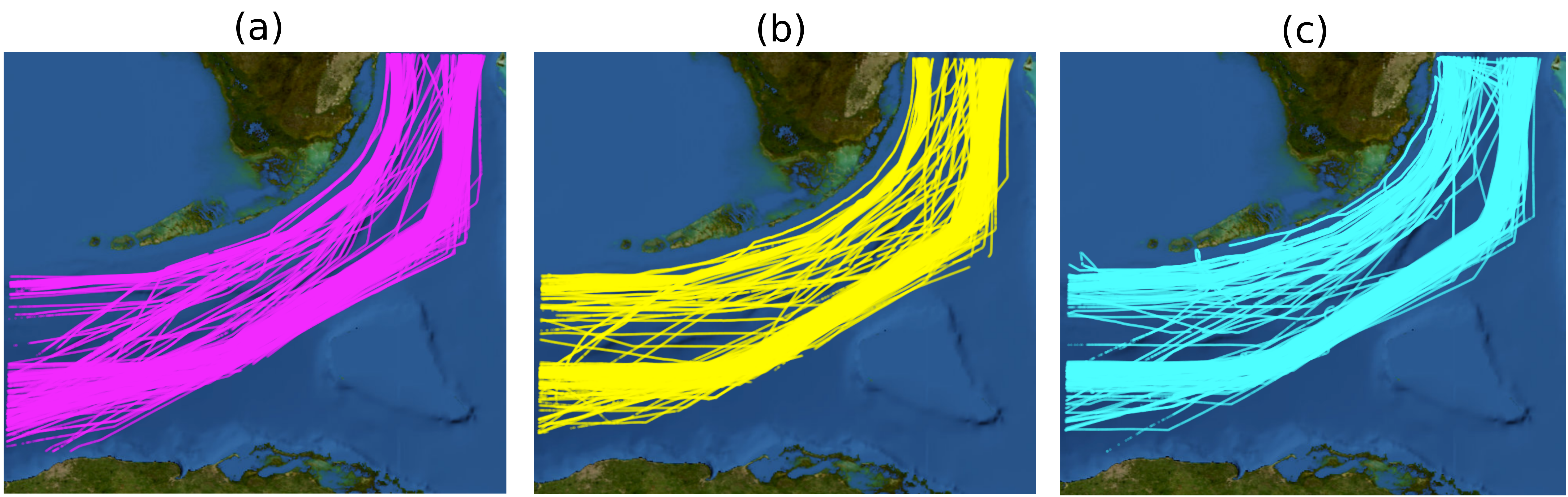}
 \caption{Vessel traffic of interest between the Miami surroundings (north east) and the entrance to the gulf of Mexico (south west) in 2015. (a) trajectories always going up (b) trajectories always going down (c) rest of the trajectories. }
 \label{fig:ais}
\end{figure}

We consider vessel traffic between the surroundings of Miami and the entrance to the gulf of Mexico in 2015 (UTM zone 17). The dataset for this is publicly available, \footnote{\url{https://marinecadastre.gov/AIS/}.} the variables of interest being the longitude ($\lambda$), the latitude ($\varphi$), the speed over ground (SOG), the course over ground (COG), and the  timestamp of every AIS signal reported by vessels moving with SOG $>$ 3 knots. We filter out some of the trajectories going to or coming from the eastern Greater Antilles as well as on the north of the Florida Keys. From the reminder, we keep trajectories which have at least 900 timesteps, leaving a dataset of about 3.2M records extracted from a totality of about 36.5M records. This is clustered into 3 categories: trajectories always going \emph{up} ($-90^{\circ}\le \textrm{COG}\le90^{\circ}$), trajectories always going \emph{down}, and the rest of the trajectories in \emph{other}. These are partially shown in Fig. \ref{fig:ais}, after removal of outliers \shortcite{Young2017} and segmentation at stop points \shortcite{Bay2017EvaluationOF} of more than an hour. They fit inside a region of interest (ROI) whose boundary is clearly appreciable from Fig. \ref{fig:ais}.    

\subsection{Partial knowledge about the update rule}
The motion of vessels is spatially-unconstrained, giving rise to maneouvers that are not seen on road traffic. Moreover, it occurs on a curved geometry and the reported $dt$ is stochastic \shortcite{ais_2014}. Nevertheless, it is our general goal to include a partial knowledge of the update rule into the models.
This is obtained by approximating the motion on the surface of the Earth as (see appendix  \ref{subsec:evol_curved} for details)
\begin{equation}\label{eq:onearth}
 \begin{split}
  \varphi_{t+1} &=\varphi_{t}+ \tfrac{1}{60}\cos(\textrm{COG}_t)\,\textrm{SOG}_t\,dt,\\
   \lambda_{t+1} &= \lambda_{t}+\tfrac{1}{60}\sin(\textrm{COG}_t)\,\textrm{SOG}_t\,dt\,/\cos(\varphi_t),\\
 \end{split}
\end{equation}
provided $dt$ (measured in hours) is small enough. Defining the state as $s_t=(\lambda_t, \varphi_t)$ and the actions targeted by the agent as $a_t=(\textrm{SOG}_t, \textrm{COG}_t, dt)$, the equations in \eqref{eq:onearth} define the functions $g_{\lambda}$ and $g_{\varphi}$ representing the partial knowledge that the agent has about the evolution of the state to be imitated. Compared to the synthetic experiments, here the agent has the extra task to learn the distribution of $dt$ for the next AIS record of a vessel to arrive. 

\subsection{Detection of abnormal motion patterns}
We train DATI with the same architecture and hyperparameters as in the synthetic experiments (except that $\pi_{\theta}^{\rarrow}$ now has three instead of two outputs). To define the train and test sets, we notice that the cluster of \emph{up} trajectories has 170 in total, the cluster of \emph{down} trajectories has 1973 in total, and the cluster of \emph{other} trajectories has 887 in total. By definition, the cluster \emph{other} has trajectories which either go up or down but not monotonically, therefore having room for exotic vessel maneouvers. To discover abnormal motion, a subset of this cluster is then used as test set.

In order to have training conditions as in the synthetic experiments, we target a similar number of training episodes. For this reason, we take the whole cluster of \emph{up} trajectories, giving 170 episodes, and downsample the cluster of \emph{down} trajectories to also have 170 trajectories (the same is done for testing on the \emph{other} cluster). These are the numbers shown in Fig. \ref{fig:ais}. Furthermore, to avoid the enconding of \emph{up} and \emph{down} as an extra feature of the state space, two DATI instances are trained, one for each cluster type.

The (exponentially-smoothed) dynamic time warping distance $D_{\textrm{dtw}}(\hat{s}_{0:T}^\theta, s_{0:T}^*)$ is now measured with respect to the great-circle distance on the surface of the Earth (unlike the Euclidean distance used for the synthetic experiments). Since the trajectories chosen for training comprise more than 300 km in length, we choose a relatively small $\varepsilon = \,$500 m to reward DATI with $r_t=\pm1$ according to whether $D_{\textrm{dtw}}(\hat{s}_{0:t}^\theta, s_{0:t})<\varepsilon$ or not.
\begin{figure}[t]
 \centering
 \includegraphics[scale=0.6]{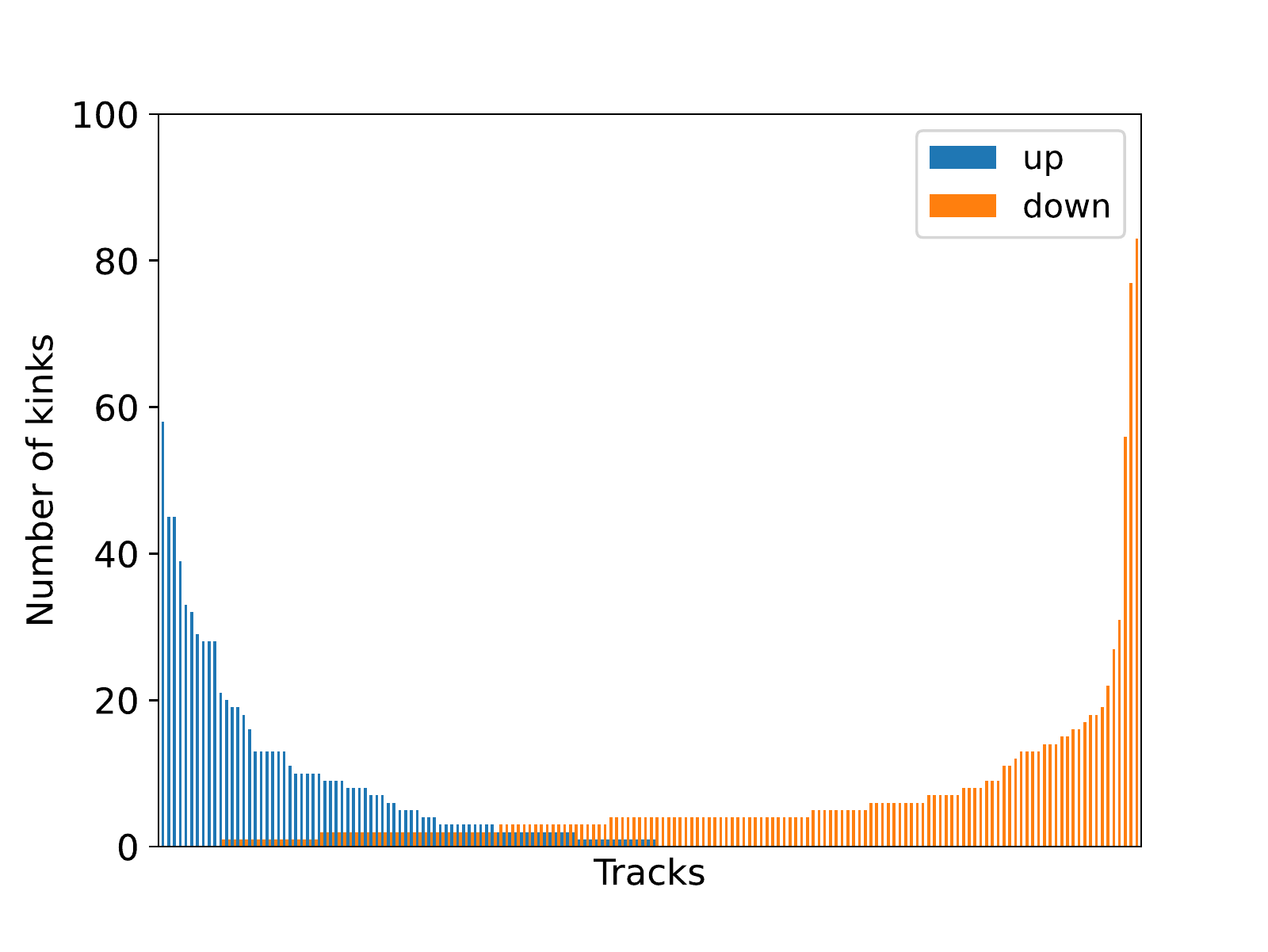}
 \caption{Number of course changes with $\Delta\textrm{COG}>10^{\circ}$ (kinks) per track in the cluster \emph{up}  and in a uniformly sampled subset of 170 trajectories from the cluster \emph{down}.}
 \label{fig:kinks}
\end{figure}

\textbf{Probing the generated distribution}. To make sense of how DATI conceives the data distribution, it is helpful to think of the main features that make up a vessel trajectory. Since there is an incentive to take the safest and shortest navigable route, trajectories may display many straight-line segments. We adopt a threshold for a significant change of course to be $\Delta\textrm{COG}=10^{\circ}$, which we call a \emph{kink}. Fig. \ref{fig:kinks} shows the amount of kinks in the train sets of both clusters. It is observed that the original cluster \emph{down} is highly skewed toward trajectories with few kinks. The distribution is multimodal, with only $7\%$ of the cluster not containing kinks, and $63\%$ of the trajectories having from 1 to 4 kinks, in the proportion of $12\%$, $19\%$, $18\%$, and $13\%$, respectively. In contrast, $49\%$ of the cluster \emph{up} does not feature any kink, which means that the trajectories in this cluster are often smooth. It is then expected that these observations are reflected in the inductive biases of the learned models.

\begin{figure}[t]
 \centering
 \includegraphics[scale=0.25]{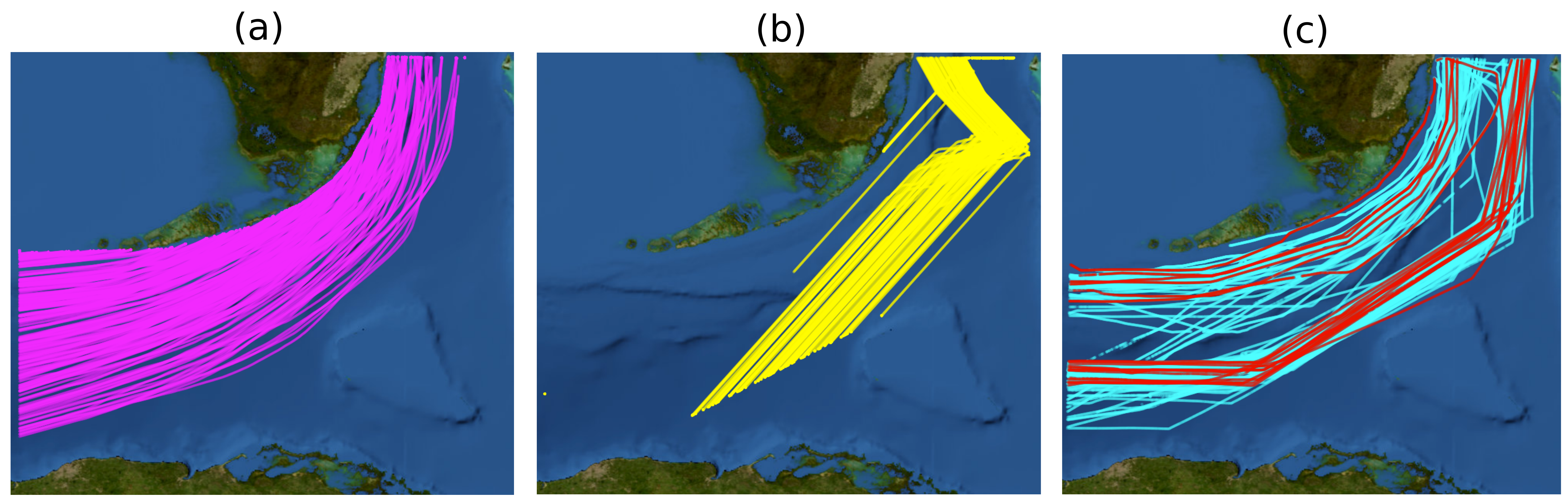}
 \caption{Trajectories generated by DATI after randomly sampling 170 initial states: (a) using the model trained with the cluster \emph{up} (b) using the model trained with the cluster \emph{down}. (c) Performance on the test set of trajectories that go up in the cluster \emph{other}, showing corresponding anomalies identified by DATI (in red). }
 \label{fig:ais_gen}
\end{figure}

We rollout the learned policies starting from 170 random points along the vertical boundary at the entrance to the gulf of Mexico for the cluster \emph{up} and along the horizontal boundary near Miami for the cluster \emph{down}. The results are shown in Fig. \ref{fig:ais_gen}(a)-(b). As expected, DATI learns to smoothly generate trajectories always going up, with shapes resambling the reference trajectories during training. It does so even in starting regions not observed during training, as can be seen by comparaing Figs. \ref{fig:ais}(a) and \ref{fig:ais_gen}(a). On the other hand, the model trained with the downsampled cluster of \emph{down} trajectories learns to generate mainly 2 kinks from the most populated mode --- this is different from the mode collapse sometimes encountered in GANs \shortcite{Durall2021CombatingMC}. However, in the attempt to fit the other types of trajectories, the end result does not resemble the shape of the trajectories observed during training, as seen by comparing Figs. \ref{fig:ais}(b) and \ref{fig:ais_gen}(b). A solution for this may be found in downsampling this cluster not uniformly but by extracting much more smooth trajectories than with kinks.   

An immediate observation from the generated trajectories is a sticking effect when the actors hit the boundaries of the ROI. This (i.e. $\hat{s}_{t+1}\rightarrow \hat{s}_t$ if $\hat{s}_{t+1}\notin\,\textrm{ROI}$) happens until the agent decides to continue exploring the interior of the ROI. It is intentionally implemented as termination condition of the episodes in all the experiments of this paper. This ensures the same number of timesteps for the generated and reference tracks. By having a lower (big) bound on the number of timesteps ($>900$) per episode, the bias from having variable horizon environments \shortcite{Kostrikov2019} is then alleviated.

The state space of DATI may easily be enlarged to include more information such as destination of the trajectories --- fixed in this work by the nature of the dataset. With this in mind, it could be used as a method of pathfinding for ocean voyages, given enough reference trajectories between source and destination. The advantage over modern methods which optimize for the shortest route \shortcite{spherical_anya} is that DATI is data-driven, and therefore can learn (as part of the distribution) highly dynamic shipping patterns often encountered in reality, which  may deviate from the shortest route, due to many varying external factors \shortcite{graphAIS}. 

\textbf{From normal to abnormal motion patterns}. We leave for a different paper how to deal with multimodal distributions of trajectories. For the proof of concept of this work, we then restrict to the analysis of abnormal patterns in the 143 trajectories $s_{0:T}^*$ belonging to the cluster \emph{other}, which start near the entrace to the golf of Mexico. DATI is run for each corresponding initial state, and $D_{\textrm{dtw}}(\hat{s}_{0:T}^\theta, s_{0:T}^*)$ is calculated and normalized with respect to the maximum value. We then ask for a threshold $\Lambda$ for which about $10\%$ of the trajectories are tagged as abnormal by setting $D_{\textrm{dtw}}(\hat{s}_{0:T}^\theta, s_{0:T}^*)>\Lambda$. This is obtained to be $\Lambda=0.75$, and the corresponding trajectories are shown in red in Fig. \ref{fig:ais_gen}(c). Independently, about $10\%$ of the most salient anomalies in the test set are manually annotated and confronted with the predictions by DATI, resulting in a weighted F1-score of $0.78$. This is very promising, given that the DATI architecture was chosen to optimize the synthetic experiments.

\section{Conclusion and outlook}
We have presented a novel method to learn representative trajectories of dynamical systems for which partial information of their update rule is known. The method does not make any assumption regarding the state representation, making it very appealing for knowledge discovery across a wide range of applications. We have demonstrated this by learning to generate representative trajectories in maritime traffic (with corresponding anomaly detection) with the same model architecture and hyperparameters with which we benchmarked the performance on synthetic datasets. Future research avenues within the trajectory data mining field include (but are not limited to) the detection of abnormal motion in more complex maritime scenarios with multi-modal distributions, in road and air traffic, pedestrian dynamics, etc. By providing a reinforcement learning environment capable of representing any family of trajectories, we encourage the research community to use a standard benchmark for trajectory imitation tasks. 

\appendix
\renewcommand{\thesubsection}{A.\arabic{subsection}}
\section*{Appendix A.}

\subsection{Shape parameters of synthetic families}
\label{subsec:shape_param}

For all families of trajectories we take $T=2\pi/\omega$. The parameter set for each family is as follows. \emph{FixedStart} has $\bm{\alpha}=\{\alpha, \omega, \eta\}$ for which $\omega=\eta=0.9$ are fixed and $\alpha$ is sampled uniformly in $[5,10]$ for each episode. \emph{UShaped} has $\bm{\alpha}=\{\alpha, \omega\}$ for which $\omega=0.9$ is fixed and $\alpha$ is sampled uniformly in $[0.2,0.8]$ for each episode. \emph{Circles} has $\bm{\alpha}=\{\alpha, \omega\}$ for which $\omega=0.4$ is fixed and $\alpha$ is sampled uniformly in $[0.5,1.0]$ for each episode. \emph{Ribbons} has $\bm{\alpha}=\{\alpha, \omega, R_1, R_2\}$ for which $\omega=0.4$, $R_1=1$, $R_2=2$ are fixed and $\alpha$ is sampled uniformly in $[-\pi, \pi]$ for each episode. 

\subsection{Hyperparameters for the synthetic experiments}
\label{subsec:hyper-param}
DATI updates the critic networks 5 times before updating the actor networks per training batch. The learning rates for the actors is $10^{-4}$, and for the critics $10^{-5}$. They are optimized using Adam with $\beta_1=0.5$ and $\beta_2=0.9$. All $L_1$ losses are optimized with a learning rate of $10^{-3}$ and weighted (with respect to the total loss) with a coefficient of $10$. The 1-Lipschitz condition for the critics is achieved by gradient penalty \shortcite{improved_wgan} with $\lambda=10$. Finally, the noise $\eta_t$ is chosen as the best between Ornstein-Uhlenbeck or Gaussian (with $\mu=0$ and $\sigma=0.3$). DDPG-TI optimizes the actor and critic networks using Adam with learning rates $10^{-4}$ and $2\times10^{-4}$, respectively. The discount factor is $\gamma=0.9$ and the rate of update of the target networks is $10^{-3}$. BC optimizes the networks using Adam with a learning rate $10^{-3}$. The maximum likelihood procedure searches for a distribution with maximum entropy, the latter condition weighted with a coefficient of $10^{-3}$. Exponential smoothing of $D_{\textrm{dtw}}$ is done with a smoothing factor of $0.9$.

\subsection{Update equation for vessel motion}\label{subsec:evol_curved}
Given a path of length $d$ on the surface of the Earth, connecting the points with geographical coordinates $(\lambda_1,\varphi_1)$ and $(\lambda_2,\varphi_2)$ --- with $(\lambda, \varphi)=(\textrm{longitude},\textrm{latitude})$ --- the angle $\theta$ subtended by the path is related to the Earth radius as $\theta=d/R$. The haversine of $\theta$ is defined as $\textrm{hav}(\theta)\equiv \sin^2(\theta/2)$ and obeys 
\begin{equation}\label{eq:hav_law}
\textrm{hav}(\theta) = \textrm{hav}(\varphi_2-\varphi_1) + \cos(\varphi_1)\cos(\varphi_2)\,\textrm{hav}(\lambda_2-\lambda_1).
\end{equation}
For small travelled distances $d/R\ll1$ and $\Delta \varphi\equiv\varphi_2-\varphi_1\ll1$ and $\Delta\lambda\equiv\lambda_2-\lambda_1\ll1$. In this limit, \eqref{eq:hav_law} becomes, after Taylor expansion,
\begin{equation}
 (d/R)^2 = \Delta\varphi^2+\cos^2(\varphi_1)\Delta\lambda^2.
\end{equation}
This expreses how the curved geometry on a sphere looks locally flat (i.e. Euclidean) as long as the axis of $\lambda$ is rescaled with $\cos(\varphi_1)$. With COG representing the angle along which the vessel moves (with respect to the geographical North), and $\textrm{SOG}\,\Delta t$ the small distance travelled during $\Delta t$
\begin{equation}
 \begin{split}
  \Delta \varphi &= \tfrac{1}{60}\cos(\textrm{COG})\,\textrm{SOG}\,\Delta t\\
  \cos(\varphi_1) \Delta \lambda &= \tfrac{1}{60}\sin(\textrm{COG})\,\textrm{SOG}\,\Delta t,\\
 \end{split}
\end{equation}
where the factor $\tfrac{1}{60}$ is used to convert from knots to degrees: $[\textrm{SOG}]=1\,\textrm{knot}=1\,\textrm{nmi}/\textrm{hour}$ and $60\, \textrm{nmi} \sim 1^{\circ}$ of longitude / latitude (i.e. The equatorial earth radius is $R=6378.137\,$km so, with $\theta=\pi\,\textrm{rad}/180$, $d=6378.14*\pi/180\,\textrm{km}=111.319\,\textrm{km}=\underline{60.1\,\textrm{nmi}}$. On the other hand, the polar radius is $R=6356.752\,$km so $d=6356.752*\pi/180\,\textrm{km}=110.946\,\textrm{km}=\underline{59.9\,\textrm{nmi}}$).

\vskip 0.2in
\bibliography{dati}

\begin{thebibliography}{}

\bibitem[\protect\BCAY{Agarwal, Schwarzer, Castro, Courville,\ \BBA\
  Bellemare}{Agarwal et~al.}{2021}]{agarwal2021deep}
Agarwal, R., Schwarzer, M., Castro, P.~S., Courville, A., \BBA\ Bellemare,
  M.~G. \BBOP2021\BBCP.
\newblock \BBOQ Deep reinforcement learning at the edge of the statistical
  precipice\BBCQ\
\newblock In {\Bem Thirty-Fifth Conference on Neural Information Processing
  Systems}.

\bibitem[\protect\BCAY{Arjovsky, Chintala,\ \BBA\ Bottou}{Arjovsky
  et~al.}{2017}]{wassersteinGAN}
Arjovsky, M., Chintala, S., \BBA\ Bottou, L. \BBOP2017\BBCP.
\newblock \BBOQ {W}asserstein generative adversarial networks\BBCQ\
\newblock In {\Bem Proceedings of the 34th International Conference on Machine
  Learning}, \BPGS\ 214--223.

\bibitem[\protect\BCAY{Arulkumaran, Deisenroth, Brundage,\ \BBA\
  Bharath}{Arulkumaran et~al.}{2017}]{Arulkumaran2017}
Arulkumaran, K., Deisenroth, M.~P., Brundage, M., \BBA\ Bharath, A.~A.
  \BBOP2017\BBCP.
\newblock \BBOQ Deep reinforcement learning: A brief survey\BBCQ\
\newblock {\Bem IEEE Signal Processing Magazine}, {\Bem 34\/}(6), 26--38.

\bibitem[\protect\BCAY{Bay}{Bay}{2017}]{Bay2017EvaluationOF}
Bay, S.~M. \BBOP2017\BBCP.
\newblock \BBOQ Evaluation of factors on the patterns of ship movement and
  predictability of future ship location in the gulf of mexico\BBCQ.

\bibitem[\protect\BCAY{Becker{-}Ehmck, Karl, Peters,\ \BBA\ van~der
  Smagt}{Becker{-}Ehmck et~al.}{2020}]{learn2fly}
Becker{-}Ehmck, P., Karl, M., Peters, J., \BBA\ van~der Smagt, P.
  \BBOP2020\BBCP.
\newblock \BBOQ Learning to fly via deep model-based reinforcement
  learning\BBCQ\
\newblock {\Bem CoRR}, {\Bem abs/2003.08876}.

\bibitem[\protect\BCAY{Belhadi, Djenouri, Lin,\ \BBA\ Cano}{Belhadi
  et~al.}{2020}]{Belhadi2020}
Belhadi, A., Djenouri, Y., Lin, J. C.-W., \BBA\ Cano, A. \BBOP2020\BBCP.
\newblock \BBOQ Trajectory outlier detection: Algorithms, taxonomies,
  evaluation, and open challenges\BBCQ.
\newblock {\Bem 11\/}(3).

\bibitem[\protect\BCAY{Brockman, Cheung, Pettersson, Schneider, Schulman,
  Tang,\ \BBA\ Zaremba}{Brockman et~al.}{2016}]{brockman2016openai}
Brockman, G., Cheung, V., Pettersson, L., Schneider, J., Schulman, J., Tang,
  J., \BBA\ Zaremba, W. \BBOP2016\BBCP.
\newblock \BBOQ Openai gym\BBCQ\
\newblock {\Bem arXiv preprint arXiv:1606.01540}.

\bibitem[\protect\BCAY{Chen\ \BBA\ Lu}{Chen\ \BBA\ Lu}{2021}]{CHEN202162}
Chen, P.\BBACOMMA\  \BBA\ Lu, W. \BBOP2021\BBCP.
\newblock \BBOQ Deep reinforcement learning based moving object grasping\BBCQ\
\newblock {\Bem Information Sciences}, {\Bem 565}, 62--76.

\bibitem[\protect\BCAY{Choi, Kim,\ \BBA\ Jin~Kim}{Choi et~al.}{2017}]{Choi2017}
Choi, S., Kim, S., \BBA\ Jin~Kim, H. \BBOP2017\BBCP.
\newblock \BBOQ Inverse reinforcement learning control for trajectory tracking
  of a multirotor uav\BBCQ\
\newblock {\Bem International Journal of Control, Automation and Systems},
  {\Bem 15\/}(4), 1826--1834.

\bibitem[\protect\BCAY{Durall, Chatzimichailidis, Labus,\ \BBA\ Keuper}{Durall
  et~al.}{2021}]{Durall2021CombatingMC}
Durall, R., Chatzimichailidis, A., Labus, P., \BBA\ Keuper, J. \BBOP2021\BBCP.
\newblock \BBOQ Combating mode collapse in gan training: An empirical analysis
  using hessian eigenvalues\BBCQ\
\newblock In {\Bem VISIGRAPP}.

\bibitem[\protect\BCAY{Freeman, Merriman, Beaver,\ \BBA\ Mueen}{Freeman
  et~al.}{2022}]{tseries2021}
Freeman, C., Merriman, J., Beaver, I., \BBA\ Mueen, A. \BBOP2022\BBCP.
\newblock \BBOQ Experimental comparison and survey of twelve time series
  anomaly detection algorithms\BBCQ\
\newblock {\Bem J. Artif. Int. Res.}, {\Bem 72}, 849–899.

\bibitem[\protect\BCAY{Giannotti, Nanni,\ \BBA\ Pedreschi}{Giannotti
  et~al.}{2006}]{Giannotti06}
Giannotti, F., Nanni, M., \BBA\ Pedreschi, D. \BBOP2006\BBCP.
\newblock \BBOQ Efficient mining of temporally annotated sequences\BBCQ\
\newblock In {\Bem In Proceedings of the 6th SIAM International Conference on
  Data Mining}, \BPGS\ 346--357.

\bibitem[\protect\BCAY{Giannotti, Nanni, Pinelli,\ \BBA\ Pedreschi}{Giannotti
  et~al.}{2007}]{Giannotti07}
Giannotti, F., Nanni, M., Pinelli, F., \BBA\ Pedreschi, D. \BBOP2007\BBCP.
\newblock \BBOQ Trajectory pattern mining\BBCQ.
\newblock KDD '07, \BPG\ 330–339, New York, NY, USA. Association for
  Computing Machinery.

\bibitem[\protect\BCAY{Goodfellow}{Goodfellow}{2017}]{Goodfellow17}
Goodfellow, I.~J. \BBOP2017\BBCP.
\newblock \BBOQ {NIPS} 2016 tutorial: Generative adversarial networks\BBCQ\
\newblock {\Bem CoRR}, {\Bem abs/1701.00160}.

\bibitem[\protect\BCAY{Gulrajani, Ahmed, Arjovsky, Dumoulin,\ \BBA\
  Courville}{Gulrajani et~al.}{2017}]{improved_wgan}
Gulrajani, I., Ahmed, F., Arjovsky, M., Dumoulin, V., \BBA\ Courville, A.~C.
  \BBOP2017\BBCP.
\newblock \BBOQ Improved training of wasserstein gans\BBCQ\
\newblock In {\Bem Advances in Neural Information Processing Systems},
  \lowercase{\BVOL}~30.

\bibitem[\protect\BCAY{Haarnoja, Zhou, Ha, Tan, Tucker,\ \BBA\ Levine}{Haarnoja
  et~al.}{2019}]{LearnToWalk19}
Haarnoja, T., Zhou, A., Ha, S., Tan, J., Tucker, G., \BBA\ Levine, S.
  \BBOP2019\BBCP.
\newblock \BBOQ Learning to walk via deep reinforcement learning\BBCQ\
\newblock In {\Bem Robotics: Science and Systems}.

\bibitem[\protect\BCAY{Hafner, Hertweck, Kloppner, Bloesch, Neunert, Wulfmeier,
  Tunyasuvunakool, Heess,\ \BBA\ Riedmiller}{Hafner et~al.}{2020}]{core_skills}
Hafner, R., Hertweck, T., Kloppner, P., Bloesch, M., Neunert, M., Wulfmeier,
  M., Tunyasuvunakool, S., Heess, N. M.~O., \BBA\ Riedmiller, M.~A.
  \BBOP2020\BBCP.
\newblock \BBOQ Towards general and autonomous learning of core skills: A case
  study in locomotion\BBCQ\
\newblock In {\Bem CoRL}.

\bibitem[\protect\BCAY{Kazemi, Goel, Eghbali, Ramanan, Sahota, Thakur, Wu,
  Smyth, Poupart,\ \BBA\ Brubaker}{Kazemi et~al.}{2019}]{t2v}
Kazemi, S.~M., Goel, R., Eghbali, S., Ramanan, J., Sahota, J., Thakur, S., Wu,
  S., Smyth, C., Poupart, P., \BBA\ Brubaker, M. \BBOP2019\BBCP.
\newblock \BBOQ Time2vec: Learning a vector representation of time\BBCQ\
\newblock {\Bem CoRR}, {\Bem abs/1907.05321}.

\bibitem[\protect\BCAY{Kim, Gu, Song, Zhao,\ \BBA\ Ermon}{Kim
  et~al.}{2020}]{dail2020}
Kim, K., Gu, Y., Song, J., Zhao, S., \BBA\ Ermon, S. \BBOP2020\BBCP.
\newblock \BBOQ Domain adaptive imitation learning\BBCQ\
\newblock In III, H.~D.\BBACOMMA\  \BBA\ Singh, A.\BEDS, {\Bem Proceedings of
  the 37th International Conference on Machine Learning}, \lowercase{\BVOL}\
  119 of {\Bem Proceedings of Machine Learning Research}, \BPGS\ 5286--5295.
  PMLR.

\bibitem[\protect\BCAY{Kontopoulos, Makris, Zissis,\ \BBA\ Tserpes}{Kontopoulos
  et~al.}{2021}]{VisionTrackClassify}
Kontopoulos, I., Makris, A., Zissis, D., \BBA\ Tserpes, K. \BBOP2021\BBCP.
\newblock \BBOQ A computer vision approach for trajectory classification\BBCQ\
\newblock In {\Bem 2021 22nd IEEE International Conference on Mobile Data
  Management (MDM)}, \BPGS\ 163--168.

\bibitem[\protect\BCAY{Kostrikov, Agrawal, Dwibedi, Levine,\ \BBA\
  Tompson}{Kostrikov et~al.}{2019}]{Kostrikov2019}
Kostrikov, I., Agrawal, K.~K., Dwibedi, D., Levine, S., \BBA\ Tompson, J.
  \BBOP2019\BBCP.
\newblock \BBOQ Discriminator-actor-critic: Addressing sample inefficiency and
  reward bias in adversarial imitation learning\BBCQ\
\newblock In {\Bem ICLR}.

\bibitem[\protect\BCAY{Last, Bahlke, Hering-Bertram,\ \BBA\ Linsen}{Last
  et~al.}{2014}]{ais_2014}
Last, P., Bahlke, C., Hering-Bertram, M., \BBA\ Linsen, L. \BBOP2014\BBCP.
\newblock \BBOQ Comprehensive analysis of automatic identification system (ais)
  data in regard to vessel movement prediction\BBCQ\
\newblock {\Bem Journal of Navigation}, {\Bem 67\/}(5), 791–809.

\bibitem[\protect\BCAY{Lazaridis, Fachantidis,\ \BBA\ Vlahavas}{Lazaridis
  et~al.}{2020}]{Lazaridis2020DeepRL}
Lazaridis, A., Fachantidis, A., \BBA\ Vlahavas, I.~P. \BBOP2020\BBCP.
\newblock \BBOQ Deep reinforcement learning: A state-of-the-art
  walkthrough\BBCQ\
\newblock {\Bem J. Artif. Intell. Res.}, {\Bem 69}, 1421--1471.

\bibitem[\protect\BCAY{Lee\ \BBA\ Kim}{Lee\ \BBA\ Kim}{2017}]{Lee2017}
Lee, H.\BBACOMMA\  \BBA\ Kim, H.~J. \BBOP2017\BBCP.
\newblock \BBOQ Trajectory tracking control of multirotors from modelling to
  experiments: A survey\BBCQ\
\newblock {\Bem International Journal of Control, Automation and Systems},
  {\Bem 15\/}(1), 281--292.

\bibitem[\protect\BCAY{Lee, Park, Lee,\ \BBA\ Lee}{Lee
  et~al.}{2019}]{sim_human_muscles2019}
Lee, S., Park, M., Lee, K., \BBA\ Lee, J. \BBOP2019\BBCP.
\newblock \BBOQ Scalable muscle-actuated human simulation and control\BBCQ\
\newblock {\Bem ACM Trans. Graph.}, {\Bem 38\/}(4).

\bibitem[\protect\BCAY{Li, Ding, Han, Kays,\ \BBA\ Nye}{Li
  et~al.}{2010}]{Li2010}
Li, Z., Ding, B., Han, J., Kays, R., \BBA\ Nye, P. \BBOP2010\BBCP.
\newblock \BBOQ Mining periodic behaviors for moving objects\BBCQ.
\newblock KDD '10, \BPG\ 1099–1108, New York, NY, USA. Association for
  Computing Machinery.

\bibitem[\protect\BCAY{Li, Han, Ding,\ \BBA\ Kays}{Li et~al.}{2012}]{Li2012}
Li, Z., Han, J., Ding, B., \BBA\ Kays, R. \BBOP2012\BBCP.
\newblock \BBOQ Mining periodic behaviors of object movements for animal and
  biological sustainability studies\BBCQ\
\newblock {\Bem Data Mining and Knowledge Discovery}, {\Bem 24\/}(2), 355--386.

\bibitem[\protect\BCAY{{Lillicrap}, {Hunt}, {Pritzel}, {Heess}, {Erez},
  {Tassa}, {Silver},\ \BBA\ {Wierstra}}{{Lillicrap} et~al.}{2016}]{ddpg}
{Lillicrap}, T.~P., {Hunt}, J.~J., {Pritzel}, A., {Heess}, N., {Erez}, T.,
  {Tassa}, Y., {Silver}, D., \BBA\ {Wierstra}, D. \BBOP2016\BBCP.
\newblock \BBOQ Proceedings of the 4th internation conference of learning
  representation (iclr)\BBCQ\
\newblock In {\Bem Continuous Control with Deep Reinforcement Learning}.

\bibitem[\protect\BCAY{Liu\ \BBA\ Hodgins}{Liu\ \BBA\
  Hodgins}{2018}]{basketball}
Liu, L.\BBACOMMA\  \BBA\ Hodgins, J. \BBOP2018\BBCP.
\newblock \BBOQ Learning basketball dribbling skills using trajectory
  optimization and deep reinforcement learning\BBCQ\
\newblock {\Bem ACM Trans. Graph.}, {\Bem 37\/}(4).

\bibitem[\protect\BCAY{Meng, Yuan, Lv, Wang,\ \BBA\ Xia}{Meng
  et~al.}{2019}]{Meng2019}
Meng, F., Yuan, G., Lv, S., Wang, Z., \BBA\ Xia, S. \BBOP2019\BBCP.
\newblock \BBOQ An overview on trajectory outlier detection\BBCQ\
\newblock {\Bem Artificial Intelligence Review}, {\Bem 52\/}(4), 2437--2456.

\bibitem[\protect\BCAY{Oh\ \BBA\ Iyengar}{Oh\ \BBA\ Iyengar}{2019}]{Min2018}
Oh, M.-h.\BBACOMMA\  \BBA\ Iyengar, G. \BBOP2019\BBCP.
\newblock \BBOQ Sequential anomaly detection using inverse reinforcement
  learning\BBCQ.
\newblock KDD '19, \BPG\ 1480–1490, New York, NY, USA. Association for
  Computing Machinery.

\bibitem[\protect\BCAY{Peng, Coumans, Zhang, Lee, Tan,\ \BBA\ Levine}{Peng
  et~al.}{2020}]{RoboImitationPeng20}
Peng, X.~B., Coumans, E., Zhang, T., Lee, T.-W.~E., Tan, J., \BBA\ Levine, S.
  \BBOP2020\BBCP.
\newblock \BBOQ Learning agile robotic locomotion skills by imitating
  animals\BBCQ\
\newblock In {\Bem Robotics: Science and Systems}.

\bibitem[\protect\BCAY{Peng, Kanazawa, Malik, Abbeel,\ \BBA\ Levine}{Peng
  et~al.}{2018}]{2018-TOG-SFV}
Peng, X.~B., Kanazawa, A., Malik, J., Abbeel, P., \BBA\ Levine, S.
  \BBOP2018\BBCP.
\newblock \BBOQ Sfv: Reinforcement learning of physical skills from
  videos\BBCQ\
\newblock {\Bem ACM Trans. Graph.}, {\Bem 37\/}(6).

\bibitem[\protect\BCAY{Pomerleau}{Pomerleau}{1991}]{Pomerleau91}
Pomerleau, D.~A. \BBOP1991\BBCP.
\newblock \BBOQ Efficient training of artificial neural networks for autonomous
  navigation\BBCQ\
\newblock {\Bem Neural Computation}, {\Bem 3\/}(1), 88--97.

\bibitem[\protect\BCAY{Raychaudhuri, Paul, van Baar,\ \BBA\
  Roy{-}Chowdhury}{Raychaudhuri et~al.}{2021}]{dripta2021}
Raychaudhuri, D.~S., Paul, S., van Baar, J., \BBA\ Roy{-}Chowdhury, A.~K.
  \BBOP2021\BBCP.
\newblock \BBOQ Cross-domain imitation from observations\BBCQ\
\newblock {\Bem CoRR}, {\Bem abs/2105.10037}.

\bibitem[\protect\BCAY{Reddy, Dragan,\ \BBA\ Levine}{Reddy
  et~al.}{2020}]{Reddy2020}
Reddy, S., Dragan, A.~D., \BBA\ Levine, S. \BBOP2020\BBCP.
\newblock \BBOQ Sqil: Imitation learning via reinforcement learning with sparse
  rewards\BBCQ\
\newblock In {\Bem 8th International Conference on Learning Representations
  (ICLR)}.

\bibitem[\protect\BCAY{Roses, Kadar, Gerritsen,\ \BBA\ Rouly}{Roses
  et~al.}{2020}]{Ross2020SimulatingOM}
Roses, R., Kadar, C., Gerritsen, C., \BBA\ Rouly, O.~C. \BBOP2020\BBCP.
\newblock \BBOQ Simulating offender mobility: Modeling activity nodes from
  large-scale human activity data\BBCQ\
\newblock {\Bem J. Artif. Intell. Res.}, {\Bem 68}, 541--570.

\bibitem[\protect\BCAY{Rospotniuk\ \BBA\ Small}{Rospotniuk\ \BBA\
  Small}{2022}]{spherical_anya}
Rospotniuk, V.\BBACOMMA\  \BBA\ Small, R. \BBOP2022\BBCP.
\newblock \BBOQ Optimal any-angle pathfinding on a sphere\BBCQ\
\newblock {\Bem J. Artif. Int. Res.}, {\Bem 72}, 475–505.

\bibitem[\protect\BCAY{Ross\ \BBA\ Bagnell}{Ross\ \BBA\
  Bagnell}{2010}]{pmlr-v9-ross10a}
Ross, S.\BBACOMMA\  \BBA\ Bagnell, D. \BBOP2010\BBCP.
\newblock \BBOQ Efficient reductions for imitation learning\BBCQ\
\newblock In {\Bem Proceedings of the Thirteenth International Conference on
  Artificial Intelligence and Statistics}, \lowercase{\BVOL}~9, \BPGS\
  661--668.

\bibitem[\protect\BCAY{Rubi, Morcego,\ \BBA\ Perez}{Rubi
  et~al.}{2021a}]{Rubi2021}
Rubi, B., Morcego, B., \BBA\ Perez, R. \BBOP2021a\BBCP.
\newblock \BBOQ Deep reinforcement learning for quadrotor path following with
  adaptive velocity\BBCQ\
\newblock {\Bem Autonomous Robots}, {\Bem 45\/}(1), 119--134.

\bibitem[\protect\BCAY{Rubi, Morcego,\ \BBA\ Perez}{Rubi
  et~al.}{2021b}]{Rubi2021b}
Rubi, B., Morcego, B., \BBA\ Perez, R. \BBOP2021b\BBCP.
\newblock \BBOQ Quadrotor path following and reactive obstacle avoidance with
  deep reinforcement learning\BBCQ\
\newblock {\Bem Journal of Intelligent {\&} Robotic Systems}, {\Bem 103\/}(4),
  62.

\bibitem[\protect\BCAY{Rubi, Perez,\ \BBA\ Morcego}{Rubi
  et~al.}{2020}]{Rubi2020}
Rubi, B., Perez, R., \BBA\ Morcego, B. \BBOP2020\BBCP.
\newblock \BBOQ A survey of path following control strategies for uavs focused
  on quadrotors\BBCQ\
\newblock {\Bem Journal of Intelligent and Robotic Systems}, {\Bem 98\/}(2),
  241--265.

\bibitem[\protect\BCAY{Salvador\ \BBA\ Chan}{Salvador\ \BBA\
  Chan}{2004}]{fastDTW}
Salvador, S.\BBACOMMA\  \BBA\ Chan, P. K.-F. \BBOP2004\BBCP.
\newblock \BBOQ Fastdtw: Toward accurate dynamic time warping in linear time
  and space\BBCQ.

\bibitem[\protect\BCAY{Solano-Carrillo, Carrillo-Perez, Flenker, Steiniger,\
  \BBA\ Stoppe}{Solano-Carrillo et~al.}{2021}]{solano2021}
Solano-Carrillo, E., Carrillo-Perez, B., Flenker, T., Steiniger, Y., \BBA\
  Stoppe, J. \BBOP2021\BBCP.
\newblock \BBOQ Detection and geovisualization of abnormal vessel behavior from
  video\BBCQ\
\newblock In {\Bem 2021 IEEE International Intelligent Transportation Systems
  Conference (ITSC)}, \BPGS\ 2193--2199.

\bibitem[\protect\BCAY{Wang, Miwa,\ \BBA\ Morikawa}{Wang
  et~al.}{2020a}]{Wang2020mining}
Wang, D., Miwa, T., \BBA\ Morikawa, T. \BBOP2020a\BBCP.
\newblock \BBOQ Big trajectory data mining: A survey of methods, applications,
  and services\BBCQ.

\bibitem[\protect\BCAY{Wang, Toyer, Gleave,\ \BBA\ Emmons}{Wang
  et~al.}{2020b}]{wang2020imitation}
Wang, S., Toyer, S., Gleave, A., \BBA\ Emmons, S. \BBOP2020b\BBCP.
\newblock \BBOQ The {\tt imitation} library for imitation learning and inverse
  reinforcement learning\BBCQ\
\newblock \url{https://github.com/HumanCompatibleAI/imitation}.

\bibitem[\protect\BCAY{Xue, Kolaric, Fan, Lian, Chai,\ \BBA\ Lewis}{Xue
  et~al.}{2021a}]{Xue2021}
Xue, W., Kolaric, P., Fan, J., Lian, B., Chai, T., \BBA\ Lewis, F.~L.
  \BBOP2021a\BBCP.
\newblock \BBOQ Inverse reinforcement learning in tracking control based on
  inverse optimal control\BBCQ\
\newblock {\Bem IEEE Transactions on Cybernetics}, 1--12.

\bibitem[\protect\BCAY{Xue, Lian, Fan, Kolaric, Chai,\ \BBA\ Lewis}{Xue
  et~al.}{2021b}]{Xue2021b}
Xue, W., Lian, B., Fan, J., Kolaric, P., Chai, T., \BBA\ Lewis, F.~L.
  \BBOP2021b\BBCP.
\newblock \BBOQ Inverse reinforcement q-learning through expert imitation for
  discrete-time systems\BBCQ\
\newblock {\Bem IEEE Transactions on Neural Networks and Learning Systems},
  1--14.

\bibitem[\protect\BCAY{Young}{Young}{2017}]{Young2017}
Young, B.~L. \BBOP2017\BBCP.
\newblock \BBOQ Predicting vessel trajectories from ais data using r\BBCQ.

\bibitem[\protect\BCAY{Zheng, Verma, Zhou, Tsang,\ \BBA\ Chen}{Zheng
  et~al.}{2021}]{survey2021}
Zheng, B., Verma, S., Zhou, J., Tsang, I.~W., \BBA\ Chen, F. \BBOP2021\BBCP.
\newblock \BBOQ Imitation learning: Progress, taxonomies and
  opportunities\BBCQ\
\newblock {\Bem CoRR}, {\Bem abs/2106.12177}.

\bibitem[\protect\BCAY{Zheng}{Zheng}{2015}]{ZhengYu2015}
Zheng, Y. \BBOP2015\BBCP.
\newblock \BBOQ Trajectory data mining: An overview\BBCQ\
\newblock {\Bem ACM Trans. Intell. Syst. Technol.}, {\Bem 6\/}(3).

\bibitem[\protect\BCAY{Zhu, Park, Isola,\ \BBA\ Efros}{Zhu
  et~al.}{2017}]{cycleGAN}
Zhu, J., Park, T., Isola, P., \BBA\ Efros, A.~A. \BBOP2017\BBCP.
\newblock \BBOQ Unpaired image-to-image translation using cycle-consistent
  adversarial networks\BBCQ\
\newblock In {\Bem International Conference on Computer Vision}.

\bibitem[\protect\BCAY{Zygouras, Spiliopoulos,\ \BBA\ Zissis}{Zygouras
  et~al.}{2021}]{graphAIS}
Zygouras, N., Spiliopoulos, G., \BBA\ Zissis, D. \BBOP2021\BBCP.
\newblock \BBOQ Detecting representative trajectories from global ais
  datasets\BBCQ\
\newblock In {\Bem 2021 IEEE International Intelligent Transportation Systems
  Conference (ITSC)}, \BPGS\ 2278--2285.

\end{thebibliography}
\bibliographystyle{theapa}

\end{document}